\pdfoutput=1

\documentclass[11pt]{article}

\usepackage[final]{acl}

\usepackage{times}
\usepackage{latexsym}

\usepackage[T1]{fontenc}

\usepackage[utf8]{inputenc}

\usepackage{microtype}

\usepackage{inconsolata}

\usepackage{graphicx}

\usepackage{amsmath}
\usepackage{pifont}
\usepackage{booktabs}
\usepackage{blindtext}
\usepackage{multirow}
\usepackage[para,online,flushleft]{threeparttable}
\usepackage{algpseudocode}
\usepackage{graphicx}
\usepackage{subfigure} 
\usepackage{enumitem}
\usepackage{tcolorbox}
\newcommand{\ie}{\emph{i.e.,}\xspace}
\newcommand{\eg}{\emph{e.g.,}\xspace}
\usepackage{listings}

\usepackage{float} 
\usepackage[ruled,vlined]{algorithm2e}

\lstdefinestyle{myListingStyle} 
    {
        basicstyle = \small\ttfamily,
        breaklines = true,
        breakindent=0pt,
    }

\title{Navigating the Nuances: A Fine-grained Evaluation of \protect\\ Vision-Language Navigation}

\author{
 \textbf{Zehao Wang\textsuperscript{1}},
 \textbf{Minye Wu\textsuperscript{1}},
 \textbf{Yixin Cao\textsuperscript{4}},
 \textbf{Yubo Ma\textsuperscript{3}},
 \textbf{Meiqi Chen\textsuperscript{2}},
 \textbf{Tinne Tuytelaars\textsuperscript{1}}
\\
 \textsuperscript{1}ESAT-PSI, KU Leuven,
 \textsuperscript{2}Peking University, \\
 \textsuperscript{3}Nanyang Technological University, 
 \textsuperscript{4}Fudan University
\\
\small{\texttt{\{zehao.wang, minye.wu, tinne.tuytelaars\}@esat.kuleuven.be}},
\small{\texttt{meiqichen@stu.pku.edu.cn}},\\
\small{\texttt{yubo001@e.ntu.edu.sg}},
\small{\texttt{yxcao@fudan.edu.cn}}
}

\begin{document}
\maketitle
\begin{abstract}

This study presents a novel evaluation framework for the Vision-Language Navigation (VLN) task. It aims to diagnose current models for various instruction categories at a finer-grained level. The framework is structured around the context-free grammar (CFG) of the task. The CFG serves as the basis for the problem decomposition and the core premise of the instruction categories design. We propose a semi-automatic method for CFG construction with the help of Large-Language Models (LLMs). Then, we induct and generate data spanning five principal instruction categories (\ie direction change, landmark recognition, region recognition, vertical movement, and numerical comprehension). Our analysis of different models reveals notable performance discrepancies and recurrent issues. The stagnation of numerical comprehension, heavy selective biases over directional concepts, and other interesting findings contribute to the development of future language-guided navigation systems. The project is now available at \href{https://zehao-wang.github.io/navnuances}{https://zehao-wang.github.io/navnuances}.

\end{abstract}

\section{Introduction}
In the Vision-Language Navigation (VLN;~\citealt{anderson2018vision_r2r}) task, an agent is instructed to navigate through virtual environments by following detailed natural language instructions. This task requires an understanding of the interplay between natural language instructions, visual cues, and the sequence of actions undertaken by the agent. This capability is crucial for a wide range of robotic applications, extending from healthcare support to everyday household assistance.

\begin{figure}[t]
  \centering
  \includegraphics[width=\columnwidth]{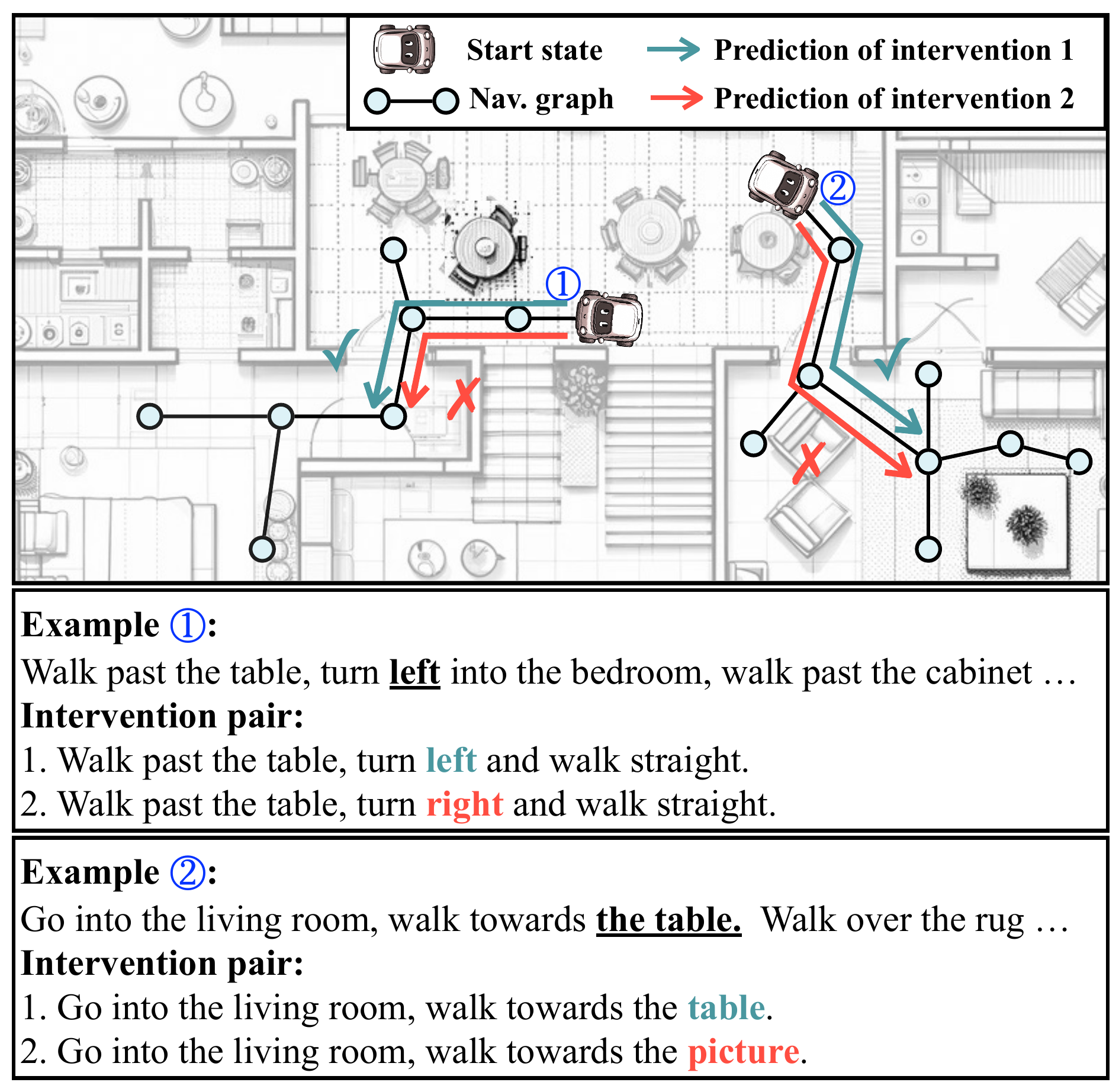}
  \caption{Examples of constructed interventions for VLN instructions. Example 1 demonstrates an intervention related to directional concepts, while Example 2 focuses on landmarks. Nonetheless, a subset of the model's predictions remains unchanged following the intervention, suggesting a deficiency in the model's ability to grasp underlying concepts. }
  \label{fig: tessar}
\end{figure}

Despite significant advancements in the latest research, we argue that the performance of VLN models may be overestimated. The current standard for evaluating vision-language navigation, as exemplified by the Room-to-Room (R2R;~\citealt{anderson2018vision_r2r}) and Room-across-Room (RxR;~\citealt{ku2020room_rxr}) datasets, predominantly hinges on endpoint success rates and broad path alignment metrics. The recent work~\cite{wang2023scaling_scalevln} suggests the performance of the state of the art is high and even quite close to human performance on these standards. Does this mean that the major challenges of the VLN task are almost solved? This perspective might be overly optimistic. For instance, a simple intervention shown in Figure.~\ref{fig: tessar} on a common VLN dataset does not trigger a consistent strong response in the model. In addition, the high success rate of a randomly navigating agent~\cite{anderson2018vision_r2r} is non-negligible. This indicates that current evaluation metrics may be insufficiently detailed. Furthermore, agents enhanced by Large Multimodal Models (LMMs;~\citealt{zhou2023navgpt,lin2024navcot}) perform unexpectedly low on standard VLN datasets. This contrasts with the strong multimodal understanding demonstrated by LMMs in other domains~\cite{fu2024mme, wake2023gpt_robot}. This discrepancy motivates us to revisit the evaluation of VLN models.

In this work, we introduce a new evaluation framework that focuses on \textit{atomic instructions}, \ie the singular actions fundamental to VLN instructions. Diagnosing VLN models at the atomic-instruction level allows us to gauge performance through various nuanced perspectives. To achieve this, we first iteratively construct a context-free grammar (CFG;~\citealt{hopcroft2001introduction_cfg}) with the help of LLMs to systematically articulate the structure of VLN task instructions. CFG, treated as a comprehensive representation of VLN instructions, allows us to induct and define atomic instruction categories. We group the components in our CFG into five main categories 
(\ie direction change, vertical movement, landmark recognition, region recognition, and numerical comprehension) and generate data accordingly to form our novel evaluation dataset \textsc{NavNuances}. For each entry in \textsc{NavNuances}, a candidate path is determined by the specific path proposing strategy according to its instruction category. The instruction is then generated using CFG and further enriched by LLMs. To ensure the data correctness, we incorporate human refinement into this automated generation process in the end. The rigorous evaluation protocols in our dataset pose significant challenges, as they require models to demonstrate a thorough understanding of individual concepts.

We benchmark various types of models based on our proposed evaluation framework. Experiments with \textsc{NavNuances} expose model discrepancies and common issues. We observe that recent advancements in the standard R2R dataset primarily stem from enhanced capabilities in vertical movement and region recognition. Despite this progress, numerical comprehension shows stagnation across various models. In terms of specific models, zero-shot agents enhanced by LLMs demonstrated even significant superiority over traditional supervised ones in handling changes in direction and recognizing landmarks. Traditional supervised approaches suffer from selective bias, often leading to deficiencies in adapting to shifts in atomic concepts, as demonstrated in Figure~\ref{fig: tessar}.

Our contributions are threefold: 
\textbf{Firstly}, we devise a comprehensive evaluation framework that addresses diverse facets of Vision-and-Language Navigation (VLN) at a granular level. 
\textbf{Secondly}, our work includes a thorough benchmarking of prevalent methodologies on ninety diverse scenes, coupled with an in-depth analysis. 
The experiments demonstrate the deficiencies and differences in the capabilities of previous models, providing valuable insights for advancing the development of VLN methods.
\textbf{Thirdly}, we present a zero-shot baseline as a minor contribution, which enhances NavGPT~\cite{zhou2023navgpt} with GPT-4-vision~\cite{achiam2023gpt4} integrating direct vision-instruction alignment.

\section{Related Work}
\subsection{Vision-language navigation Datasets}
Vision-Language Navigation (VLN;~\citealt{anderson2018vision_r2r}) tasks integrate language guidance within embodied environments. This task is initially introduced by the Room-to-Room dataset (R2R;~\citealt{anderson2018vision_r2r}) which requires step-by-step navigation in virtual spaces. Subsequent research expanded this framework through variations like multilingual RXR datasets~\cite{ku2020room_rxr} and addressed more complex navigation challenges.
The advent of conversational interfaces led to interactive VLN tasks, exemplified by CVDN~\cite{thomason2020vision_CVDN} and Teach~\cite{padmakumar2022teach}, fostering navigation via dialogue interpretation. Concurrently, efforts like VLN-CE~\cite{krantz_vlnce_2020} aimed to transition VLN tasks into continuous environments. Despite these advancements, a nuanced evaluation of VLN models on atomic-level instructions remained underexplored. Our work addresses this by developing a dataset specifically designed to assess the fundamental capabilities of VLN agents, thereby contributing to the refinement of models across various VLN settings.

\subsection{Models in VLN tasks}
The introduction of the R2R dataset~\cite{anderson2018vision_r2r} catalyzed the development of numerous models focusing on VLN tasks in discrete environments. Early efforts, such as the Seq2Seq~\cite{anderson2018vision_r2r} and RCM~\cite{wang2019reinforced_rcm} models, emphasized training strategies leveraging Imitation and Reinforcement Learning within a conventional front-view framework. Subsequent innovations like CLIP-ViL~\cite{shen2021much_clipvil} augmented these models with advanced visual features from CLIP~\cite{radford2021learning_clip}. Attention then turned to the effective encapsulation of historical data, with approaches like VLN-BERT~\cite{hong2021vlnbert} utilizing recurrent transformer structures, and HAMT~\cite{chen2021history_hamt} compactly encoding historical visual cues. More recent endeavors~\cite{chen2022think_duet,an2023bevbert} have explored the integration of topological or even metric maps to enrich navigational contexts. Parallel to these model-centric advancements, initiatives such as ScaleVLN~\cite{wang2023scaling_scalevln} aimed at scaling up training data. More recently, the research focus has switched to exploring VLN with LLMs~\cite{zhou2023navgpt,long2023discussnav,chen20232_a2nav,lin2024navcot}. Despite these significant strides, a comprehensive understanding of how these methodologies enhance specific VLN abilities, particularly atomic instruction comprehension, remains unclear. Our work seeks to shed light on this fundamental aspect and offers insights into the underlying capabilities necessary for effective VLN.

\section{NavNuances Dataset}

The challenge of curating a nuanced dataset is to comprehensively cover the atomic categories in VLN instructions.
To achieve this, our approach begins by iteratively constructing a context-free grammar (CFG) with the help of LLM to articulate and cover all components of VLN instructions in a unified representation (Section~\ref{sec: cfg}). Then, we induct and categorize the atomic components of the CFG into five principal categories (Section~\ref{sec: main cat}). Building on these categorizations, we develop a semi-automatic process for data annotation of each atomic instruction category, adhering to the CFG-defined natural instruction standards (Section~\ref{sec: collection}).

\subsection{The Context-Free Grammar for VLN}
\label{sec: cfg}

Our CFG defines a set of rules and concepts that structure the instructions in VLN. 
It can be formalized as a quadruple, \ie $\text{CFG} = (N, T, P, S)$. Non-terminals $N$ (in uppercase such as Landmark in List 1) represent broader conceptual categories or composite concepts. Terminals $T$ signify specific actionable elements or descriptors and are denoted by lowercase words (\eg left, right). Production Rules $P$ within the CFG outline how various elements are combined to form higher-level Non-terminals. And Start Symbol $S$ triggers the instruction generation process. An illustrative instruction such as \textit{walk past the red chair} can be generated by the pattern ``\textit{ActionO}+ \textit{Landmark}(\textit{Modifier(Attribute)} + \textit{Object})" in List 1. The complete version of CFG is available in the supplementary materials (Appendix~\ref{ap: cfg}). 

To ensure the integrity and completeness of the CFG, we instruct GPT-4~\cite{achiam2023gpt4} to parse the instructions in standard datasets (R2R~\cite{anderson2018vision_r2r} and RxR~\cite{ku2020room_rxr}) using the CFG and identify any omissions in the current CFG.
Through an iterative refinement process incorporating manual adjustment, we continuously update the CFG until GPT-4 can no longer detect missing components. An example is illustrated in Appendix~\ref{ap: cfg eg}. The final CFG is defined at the conceptual level and ignores linguistic variations linked to the same concept. For example, the phrases ``move towards" and ``go towards" are both represented by the same terminal ``walk towards" in CFG.

\subsection{Atomic Instruction Categories}
\label{sec: main cat}

\begin{algorithm}[t]
\caption{Context-free grammar (partial)}
\begin{algorithmic}[1]
\State $S \rightarrow Vp$
\State $Vp \rightarrow \text{ActionT}$
\State \hspace{\algorithmicindent} $\vert \text{ActionS}$
\State \hspace{\algorithmicindent} $\vert \text{ActionO} + \text{Landmark}$
\State \hspace{\algorithmicindent} $\vert \text{ActionR} + \text{Region}$
\State \hspace{\algorithmicindent} $\vert Vp + Vp$
\State \hspace{\algorithmicindent} $\vert Vp + Ir$
\State $Ir \rightarrow (\textit{action irrelevant sentence})$
\State $Numerical \rightarrow \text{first} \vert \text{second} \vert \text{third} \vert \ldots$
\State $Room \rightarrow \text{room} \vert \text{kitchen} \vert \text{bathroom} \vert \ldots$
\State $Direction \rightarrow \text{left} \vert \text{right}$
\State $Object \rightarrow \text{bed} \vert \text{table} \vert \text{chair} \vert \ldots$
\State $Attribute \rightarrow \text{red} \vert \text{yellow} \vert \ldots$
\State $Modifier \rightarrow Attribute \vert ... \vert \epsilon $
\State $Landmark \rightarrow Modifier + Object$
\State $ActionO \rightarrow \text{``walk past''} \vert \text{``walk towards''}| ...$
\State ...
\end{algorithmic}
\end{algorithm}

\begin{figure*}[t]
  \centering
  \includegraphics[width=0.98\textwidth]{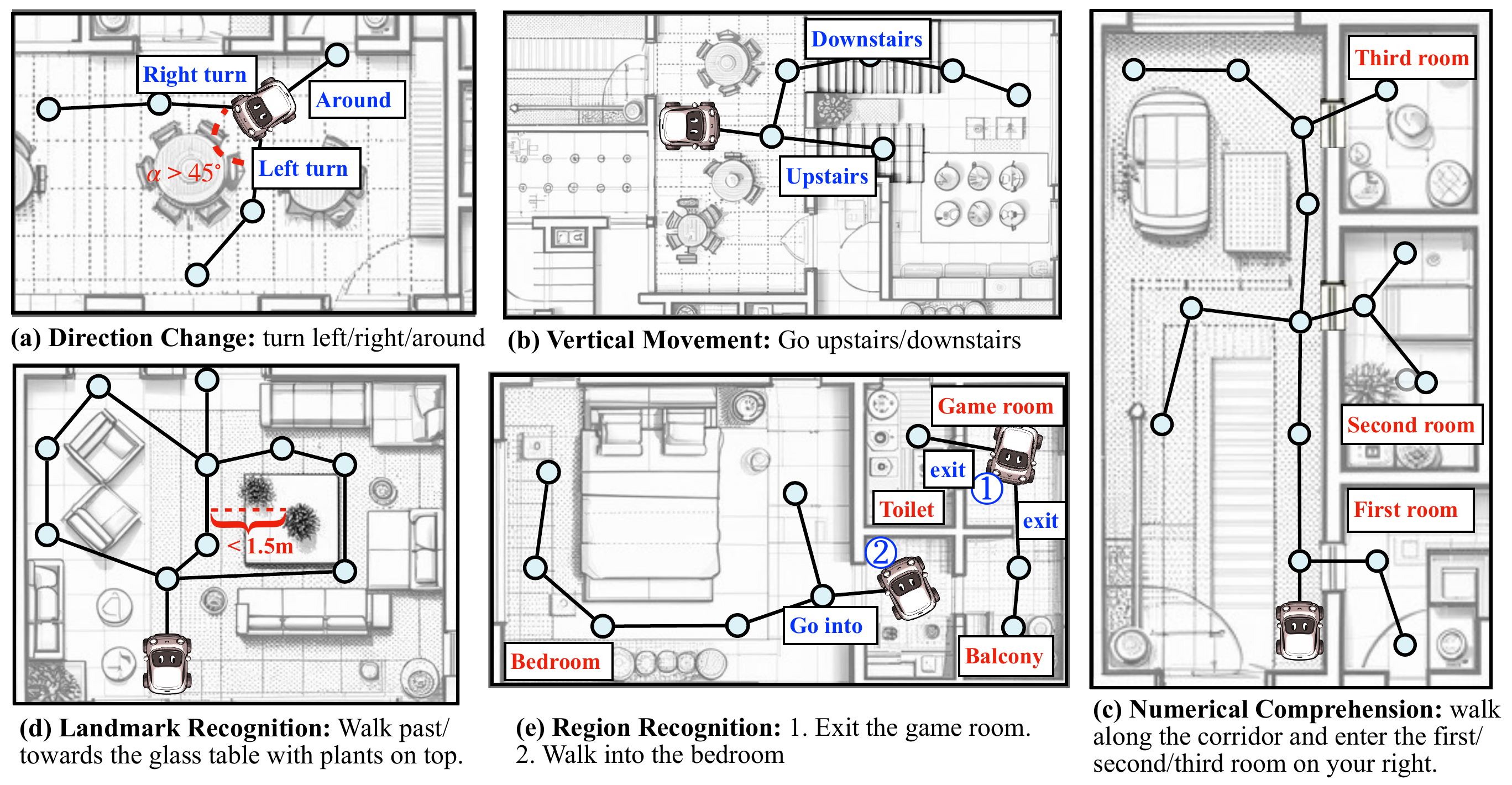}
  \caption{Schematic diagram of annotation criteria for five main categories in the \textsc{NavNuances} dataset. }
  \label{fig: method}
\end{figure*}

CFG provides a comprehensive yet elegant representation of VLN instructions. Based on this, we can discern what kind of concepts or patterns are fundamental to the VLN instructions. This further enables us to induct the atomic instruction categories. We define five primary categories introduced below:

\noindent\textbf{Direction Change:} This category stems from the CFG's \textit{ActionT}, which encapsulates turning actions. These turning actions are distinct because they exclusively pertain to directional changes and are independent of the agent's observations.

\noindent\textbf{Vertical Movement:} Derived from the \textit{ActionS}, this category is associated with movements in the vertical plane, such as ascending or descending stairs. It highlights the agent's interactions with vertical elements in the environment.

\noindent\textbf{Numerical Comprehension:} Numerical comprehension, \ie being able to count or even calculate, is quite hard yet crucial for an agent. This is challenging even for the latest LLMs~\cite{stolfo2023causal,lu2023mathvista}.  We propose to separately evaluate this category in the VLN task. It focuses on instructions that incorporate an unambiguous numerical concept, aiding in the identification of multiple landmarks or regions. 

\noindent\textbf{Landmark Recognition:} This category is inducted from production rules that involve a singular center object. It encompasses instructions that direct the agent towards or past specific landmarks within the environment.

\noindent\textbf{Region Recognition:} Similar to Landmark Recognition, this category pertains to instructions related to distinct areas or rooms. 

\subsection{Dataset Construction}
\label{sec: collection}
Our dataset is collected from 90 Matterport~\cite{chang2017matterport3d_mp3d} virtual environments aided by the semantic annotations in the Habitat simulator~\cite{szot2021habitat}. Our annotation workflow consists of four stages: rule-based candidate path proposing, CFG-driven instruction crafting, human refinement, and linguistic enrichment via rephrasing by LLMs. Each annotated datum contains a natural language instruction, the agent's initial pose, and annotations for evaluation purposes, such as the ground truth paths or landmark locations. The statistics of the \textsc{NavNuances} dataset can be found in Table~\ref{tab: stat dset}, with detailed examples provided in Appendix~\ref{ap: dataset}.

\begin{table*}[ht]
\centering
\begin{tabular}{|l|c|l|}
\hline
\textbf{Category} & \textbf{Total Instances} & \textbf{Subsets (Instances)} \\ \hline
\textbf{Direction Change (DC)} & 579 & \begin{tabular}[c]{@{}l@{}} Turn right: 192; left: 192; around: 195 \end{tabular} \\ \hline
\textbf{Vertical Movement (VM)} & 170 & \begin{tabular}[c]{@{}l@{}} Go upstairs: 87; downstairs: 83 \end{tabular} \\ \hline
\textbf{Numerical Comprehension (NU)} & 78 & \begin{tabular}[c]{@{}l@{}} 1st: 31; 2nd: 24; 3rd: 13; > 4th: 10 \end{tabular} \\ \hline
\textbf{Region Recognition (RR)} & 275 & \begin{tabular}[c]{@{}l@{}} Go into: 105; Exit: 170 \end{tabular} \\ \hline
\textbf{Landmark Recognition (LR)} & 685 & \begin{tabular}[c]{@{}l@{}} Walk towards: 353; Walk past: 332 \end{tabular} \\ \hline
\end{tabular}
\caption{Main Statistics of the \textsc{NavNuances} Dataset}
\label{tab: stat dset}
\end{table*}

\noindent\textbf{Direction Change: } Instructions in the direction change category direct the agent to make turns. We eliminate ambiguity by selecting junctions with a clear divergence in path directions (adjacent paths exceeding a large angle between them as depicted in Figure~\ref{fig: method}a), ensuring distinct navigation choices. We integrate forward movement into the instruction crafting to accommodate different VLN models and to facilitate evaluation, leading to concise instructions like \textit{``turn left/right/around, then walk straight"}. Humans are involved in refining the dataset by excluding starting positions without central obstacles in the view. This exclusion is necessary; some instances meet the selection criteria only because the navigation graph is sparse. We additionally annotate the paired instructions for left and right turns with the same starting view.

\noindent\textbf{Vertical Movement: } Vertical movement in VLN tasks is typically confined to ascending or descending stairs. Therefore, we identify the longest paths within 3D bounding boxes labeled by 'stairs' in each environment. The instruction template is straightforward containing only \textit{``go upstairs/downstairs and stop on the next floor"}. Given the bounding boxes' imprecision, human annotators are involved to adjust the start and end positions. For views that encompass two staircases in opposite vertical directions, annotators are instructed to mark these special positions and annotate paired paths from the same starting viewpoint, as shown in Figure~\ref{fig: method}b. 
This subset is small but important for assessing awareness of the vertical direction. The human-refined trajectories are considered as the ground truth and included in the dataset for evaluation purposes.

\noindent\textbf{Numerical Comprehension: } This category emphasizes the memory of sequential elements and instance-level identification. We focus on region-level numerical comprehension, utilizing the semantic annotations of 'hallway'. The process begins by filtering out hallways with insufficient number of doors and using the longest paths within to determine the starting positions. Subsequently, human annotators are asked to annotate the room count and the respective sides while navigating. The instruction follows the template: \textit{``walk along the corridor and turn into the ith room on your left/right"}. An example case is shown in Figure~\ref{fig: method}c. Paths that share identical initial poses, yet differ in numerical and directional values, are treated as negative data. These are included in the dataset to support the evaluation of numerical comprehension.

\noindent\textbf{Landmark Recognition: } This category requires taking a path associated with a specific landmark. To assess landmark recognition capabilities, it is necessary to provide instance-level descriptions in the instructions. We begin by identifying potential navigable objects using semantic annotations. We leverage GPT-4-vision~\cite{achiam2023gpt4} for precise object category identification and instance-level description generation given the view orientated towards the object. We then construct paths that meet specific criteria regarding curvature and proximity to the object's center, as shown in Figure~\ref{fig: method}d. The resulting instructions encompass actions such as \textit{``walk past + modifier + object"}. We include manual checks and modifications to ensure the visibility of target landmarks from starting viewpoints. The object center is included in the dataset as supplementary information for evaluation.

\noindent\textbf{Region Recognition: } 
Finally, the region recognition category is narrowed down to 'go into' and 'exit' actions due to the potential ambiguity in 'go through' instructions. Unlike specific endpoint-related data, region-related data pertains to a set of points associated with the concept. For example, in Figure~\ref{fig: method}e, given a starting point and the instruction \textit{``go into the bedroom"}, we record all points inside adjacent bedrooms as correct responses. For \textit{``exit the dining area"}, all areas outside the current room are marked as valid positions.

\section{Experiment}
We conduct a comprehensive evaluation of various existing VLN models across the five main categories in our \textsc{NavNuances} dataset. 
\begin{table*}[t]
\centering
\begin{threeparttable}
\centering
\small
\setlength\tabcolsep{3pt}      
    \begin{tabular}{@{}ll|ccc| ccccc | ccc } 
    \toprule      &Method & \multicolumn{3}{c|}{\textbf{Experimental setting}}  &  \multicolumn{5}{c|}{\textbf{Evaluation Results}} & \multicolumn{3}{c}{\textbf{R2R unseen}}\\
    & & Vision & Action & History & DC & NU & LR & RR & VM & SR & nDTW & SPL \\
    \midrule
     \parbox[t]{2mm}{\multirow{8}{*}{\rotatebox[origin=c]{90}{Supervised}}}
    & \shortstack{Random} & None & viewpoint  & None & 36.79 & 7.69 & 30.22 & 57.45 & 11.76 & 15.88 & 24.21 & 14.04 \\
    & \shortstack{Seq2Seq} & front-view & rule-based  & hidden state & 75.30 & 21.79 & 22.04 & 53.09 &25.88 & 21.46 & 25.04 & 18.50 \\
    & \shortstack{CLIP-ViL} & front-view & rule-based & hidden state & 77.20 & 29.49 & 36.78 & 74.18 & 69.41 & 52.15 & 47.75 & 47.64 \\
    & \shortstack{VLN-BERT} & panorama & viewpoint & hidden state & 72.02 & 29.49 & 36.05 & 80.36 & 75.29 & 62.75 & 65.49 & 56.89 \\
    & \shortstack{HAMT} & panorama & viewpoint & past views & 79.62 & 28.21 & 36.05 & 77.81 & 68.82 & 63.22 & 66.37 & 57.70 \\
    & \shortstack{DUET} & panorama & viewpoint & topo. map & 64.76 & 26.92 & 35.76 & 77.45 & 76.47 & 71.52 & 67.02 & 60.41 \\
    & \shortstack{BEVBERT} & panorama & viewpoint & topo./metric map & 63.21 & 24.35 & 30.22 & 80.36 & 84.12 & 75.18 & 69.40 & 63.68 \\
    & \shortstack{ScaleVLN} & panorama & viewpoint & topo. map & 72.88 & 26.92 & 29.92 & 84.73 & 84.71 & 80.97 & 74.76 & 70.33 \\

    \midrule
    \parbox[t]{2mm}{\multirow{3}{*}{\rotatebox[origin=c]{90}{0-shot}}}
    & \shortstack{NavGPT3.5} & pano. text & viewpoint & text history &81.87  &20.51  &58.54 &  39.63 & 7.06 & 12.67 & 40.82 & 11.45 \\
    & \shortstack{NavGPT4} & pano. text & viewpoint & text history & 91.87 & 34.78 & 54.83 & 67.61 & 11.36 & 34.78 & 47.53 & 31.64 \\
    
    & \shortstack{NavGPT4v} & panorama & viewpoint & text history & 92.68 &  39.13 & 62.87 &  56.25 & 13.64 &  41.30 & 54.78  & 36.84 \\
    \midrule
    & \shortstack{Human} & front-view & turn/vpt. & memory &95.83  & 89.13 & 89.44  & 89.89 & 94.42 &-  & - & - \\
    \bottomrule 
    \end{tabular}
    \caption{\textbf{Main Results} for baselines evaluated on five main categories of \textsc{NavNuances} dataset, \ie Direction Change (DC), Vertical Movement (VM), Landmark Recognition (LR), Region Recognition (RR) and Numerical Comprehension (NU). We also post the reproduced results on the standard R2R unseen dataset using three principal metrics: Success Rate (SR), normalized Dynamic Time Warping (nDTW) and Success rate weighted by normalized inverse Path Length (SPL) } 
    \label{tab: main} 
\end{threeparttable}
\end{table*}

\subsection{Baselines}
In this study, we examine baseline models categorized by input modalities, action spaces, memory representations, and supervision approaches. Input modalities range from front-view RGB images (\eg Seq2Seq model~\cite{anderson2018vision_r2r}) and panorama images (\eg VLN-BERT~\cite{hong2021vlnbert}) to textual descriptions of panorama views (\eg NavGPT~\cite{zhou2023navgpt}). Models differ in their action space, utilizing viewpoint selection (\eg ScaleVLN~\cite{wang2023scaling_scalevln}), predefined rule-based actions (\eg Seq2Seq~\cite{anderson2018vision_r2r}), or a combination thereof. Memory representation varies among models, employing hidden states (\eg CLIP-ViL~\cite{shen2021much_clipvil}), past visual inputs (\eg HAMT~\cite{chen2021history_hamt}), topological (\eg DUET~\cite{chen2022think_duet}) or metric maps (\eg BEVBERT~\cite{an2023bevbert}), or interactive chat histories (\eg NavGPT~\cite{zhou2023navgpt}). Except for differences in the pretraining data sources, all the supervised models are fine-tuned on the R2R dataset~\cite{anderson2018vision_r2r}. More details are available in Appendix~\ref{ap: baseline}.

We introduce \textbf{NavGPT4v}, an enhancement of the text-based NavGPT~\cite{zhou2023navgpt} model with visual inputs, integrating actual image views with GPT-4-vision~\cite{achiam2023gpt4}. We modify the initial prompt in NavGPT to highlight the presence of visual resources and their relevance to a particular direction, as illustrated in Appendix~\ref{ap: prompt}. This development targets incorporating direct visual information to capture essential details that pre-captioning might miss.

\subsection{Evaluation Protocols} 

In this section, we introduce the evaluation protocols for our Vision-Language Navigation (VLN) evaluation set. These protocols are designed to precisely measure the performance of navigation models based on detailed success criteria for different categories of atomic instructions. 

For categories \textbf{Landmark Recognition}, \textbf{Numerical Comprehension}, and \textbf{Vertical Movement}, the evaluations follow the distance-related protocols. The criteria differ slightly depending on the nature of the movement. For instance, in the vertical movement category, success is defined by a 3-meter radius to a specified endpoint. For instructions involving more localized navigation, such as walking towards a landmark, the metric focuses more on the reduction in distance to the landmark. Further details can be found in Appendix~\ref{ap: protocols}.

\textbf{Region Recognition} category is more related to inclusion-related protocol. Distance metrics are inadequate due to the lack of a precise endpoint. Success in this category is defined by the model's ability to stop within a designated region, determined by its boundaries.

For the \textbf{Direction Change} category, we evaluate the model's compliance with directional instructions. The protocol involves dividing the area around the starting point into sectors to assess the accuracy of the model's initial movement direction in response to the given instruction.

\subsection{Main Results}
We report the performance evaluated on \textsc{NavNuances} as well as the reproduced results on the unseen validation split of the R2R dataset~\cite{anderson2018vision_r2r} in Table~\ref{tab: main}. We assess NavGPT4 and NavGPT4v using a random subset of around 130 samples, ensuring replicability of the officially reported NavGPT performance without incurring significant API costs. 

Reflecting on the advancements in the standard R2R dataset, it appears that improved layout and spatial understanding underpin the progress of VLN models. This is evident from the results in \textbf{vertical movement (VM)} and \textbf{region recognition (RR)} tasks on our dataset. 
This correlation is probably due to the statistics of the R2R unseen split. We find that more than 35\% of the instructions necessitate navigation through stairs, and the majority involve concepts related to rooms. 
%
The correlation is observed consistently across different models. For instance, CLIP-ViL's leap in performance on the R2R unseen split compared to the prior model Seq2Seq (30.69\% absolute increase in success rate) correlates with significant gains in vertical movement (from 25.88\% to 69.41\%) and region recognition (from 53.09\% to 74.18\%). And the low performance of zero-shot methods on R2R also follows the lower success rates in these tasks.

Despite advancements, there is a noticeable stagnation in models' \textbf{numerical comprehension (NU)}, likely due to the sparse numerical data in existing datasets and the non-essential nature of numerical comprehension for task completion. Compared to traditional methods, LLM-enhanced models show slightly better performance but still fall significantly short of human capabilities. These findings highlight that numerical comprehension presents a substantial challenge across various model types, the inference ability w.r.t. numerical values requires further improvement.

In examining \textbf{directional changes (DC)} within supervised methods, models with explicit directional commands (such as the methods with rule-based action space, Seq2Seq, and CLIP-ViL) can easily reach or even outperform those employing viewpoint selection techniques, suggesting the importance of clear action spaces for effective turning choices. This is further supported by the superior performance of zero-shot agents, as each observation in the zero-shot agent's prompt includes a clear description of its orientation.

\begin{figure}[t]
  \centering
  \includegraphics[width=\columnwidth]{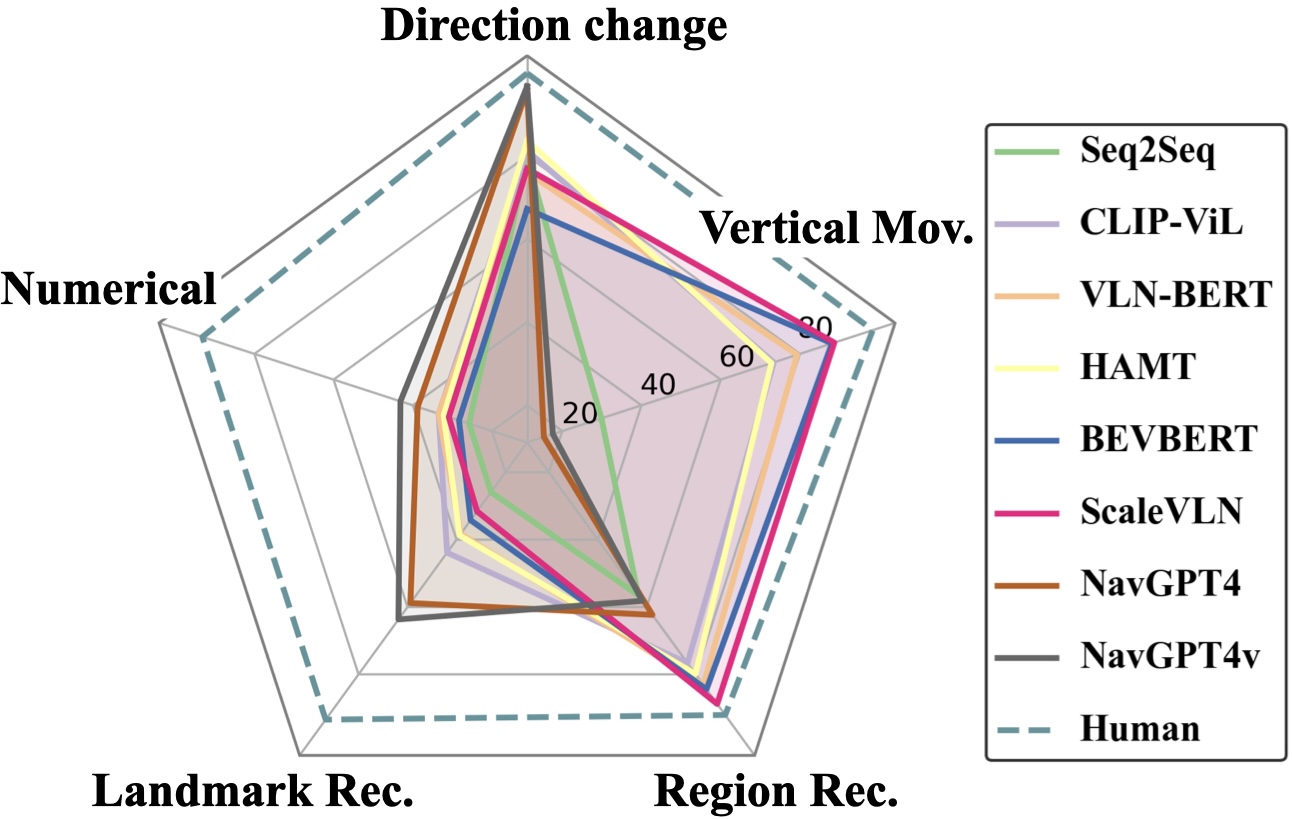}
  \caption{The success rate of models evaluated on five main categories of NavNuances dataset. Human performance is denoted by the green dashed line. }
  \label{fig: radar}
\end{figure}

In \textbf{landmark recognition (LR)}, zero-shot agents outperform models supervised on R2R data, which shows a contrast to their performance on the standard R2R dataset. This indicates that extensive knowledge of large pre-trained models can overcome the constraints inherent in small-scale supervised training. A notable comparison between NavGPT4 and our NavGPT4v reveals that conditioning observations on specific instructions leads to more accurate landmark recognition, attributed to the richness of visual content beyond mere captions. In addition, the high performance of the random agent suggests the choices within a fixed radius are limited. This highlights the limitations of supervised agents. Their performance, while comparable to the random agent, falls short of a true understanding of individual object instances.

Compared to traditional supervised methods, the performances of LMMs (\eg NavGPT4v) on \textbf{Region Recognition (RR)} and \textbf{Vertical Movement (LM)} are significantly lower. This trend is more clear as illustrated in Figure~\ref{fig: radar}. The diminished performance in vertical movement may be attributed to a lack of contextual understanding, as discussed in the recent multimodal benchmark CODIS ~\cite{luo2024codis}. Regarding the surprisingly low performance on region recognition, this issue seems to stem from the LMM's imprecise boundary judgment; the model tends to prematurely halt while merely observing the target region. Further discussion is available in Appendix~\ref{ap: room}.

\subsection{Additional Experiments}

\noindent\textbf{Does the agent understand numerical values? } 

\begin{figure}[t]
  \centering
  \includegraphics[width=\columnwidth]{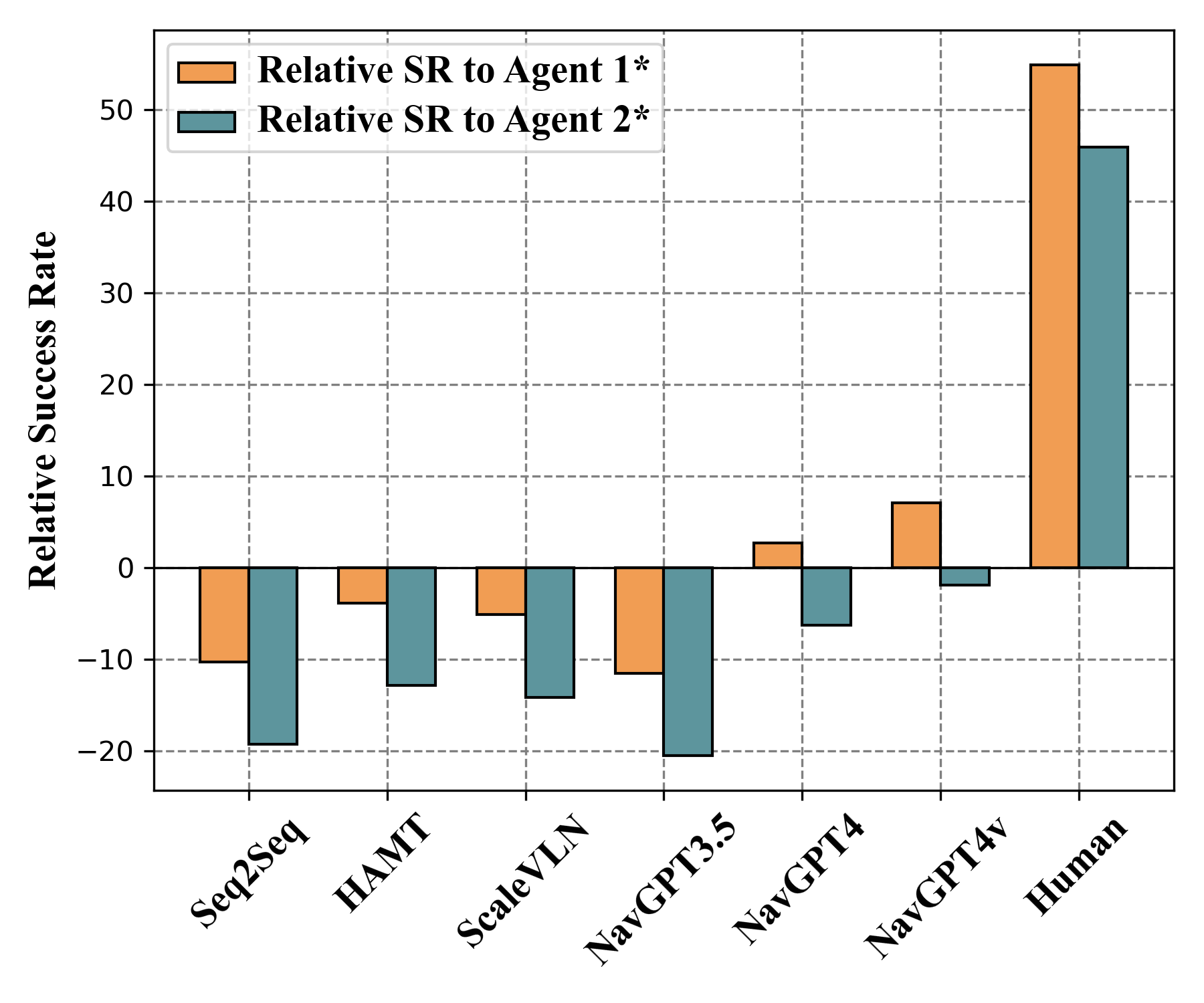}
  \caption{Success rate relative to two additional random agents in the numerical comprehension category. \textit{Agent 1*} is the random agent that knows the concept of entering the room in the corridor. \textit{Agent 2*} is the random agent which also has directional awareness. The success rates of \textit{Agent 1*} and \textit{Agent 2*} are 32.06\% and 41.03\%. }
  \label{fig: add. numerical}
\end{figure}

In this additional experiment, we aim to further study the numerical comprehension capabilities of models. Despite observing an overall low performance in this category, these models do show some improvements over a baseline random walk agent. However, the "numerical" concept is a special modifier, which always links to a specific object or region with a certain spatial relation. This association leads to an overestimation of the capability of numerical comprehension. Thus, we introduce two additional random agents to isolate these factors: The first agent simulates a basic understanding of spatial layouts (\textit{Agent 1*}), enabling the agent to select a room to enter. The second agent embodies directional intelligence (\textit{Agent 2*}), allowing the agent to choose a room on the specific side, such as entering a room on the left.

As shown in Figure~\ref{fig: add. numerical}, for some of the supervised models such as HAMT and ScaleVLN, the performance is comparable (relative success rate approaches zero) to that of the \textit{Agent 1*} but significantly lagged behind the \textit{Agent 2*}. Zero-shot agents enhanced by GPT-4 can surpass  but still have much lower performance than humans (50\% success rate below). This discrepancy highlights a critical gap in current models: while they may grasp basic layout concepts to a degree, their understanding of more complex scenarios involving both numerical values and directional cues is markedly deficient. The results, as illustrated in Figure~\ref{fig: add. numerical}, highlight the need for advanced models that integrate numerical, layout, and directional understanding.

\noindent\textbf{Can the model understand specific landmarks and the spatial relation with them? } 
\begin{figure}[t]
  \centering
  \includegraphics[width=\columnwidth]{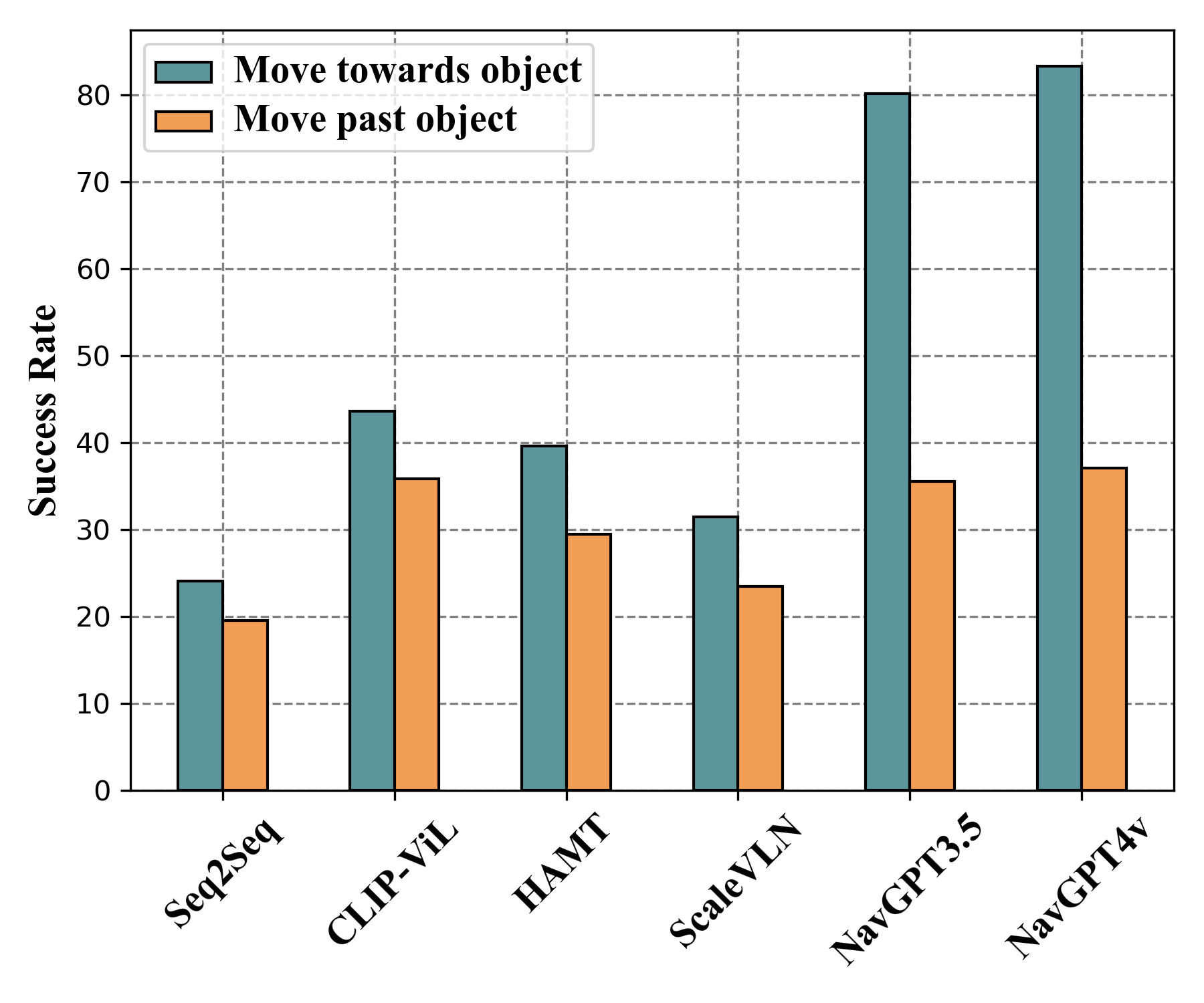}
  \caption{Results of two subsets of the landmark recognition category in the \textsc{NavNuances} dataset. The significant gap of the 'moving towards' subset comes from large pre-trained vision models since NavGPT3.5}
  \label{fig: add. difficulty in sub-category of LR}
\end{figure}

In the Landmark recognition category, we further assess the models' performance in its two distinct subsets: navigating towards a specific object and navigating past an object. The former primarily tests the models' visual grounding capabilities, while the latter introduces an additional layer of complexity by requiring an understanding of spatial relationships based on sequential observations.

\begin{figure}[t]
  \centering
  \includegraphics[width=\columnwidth]{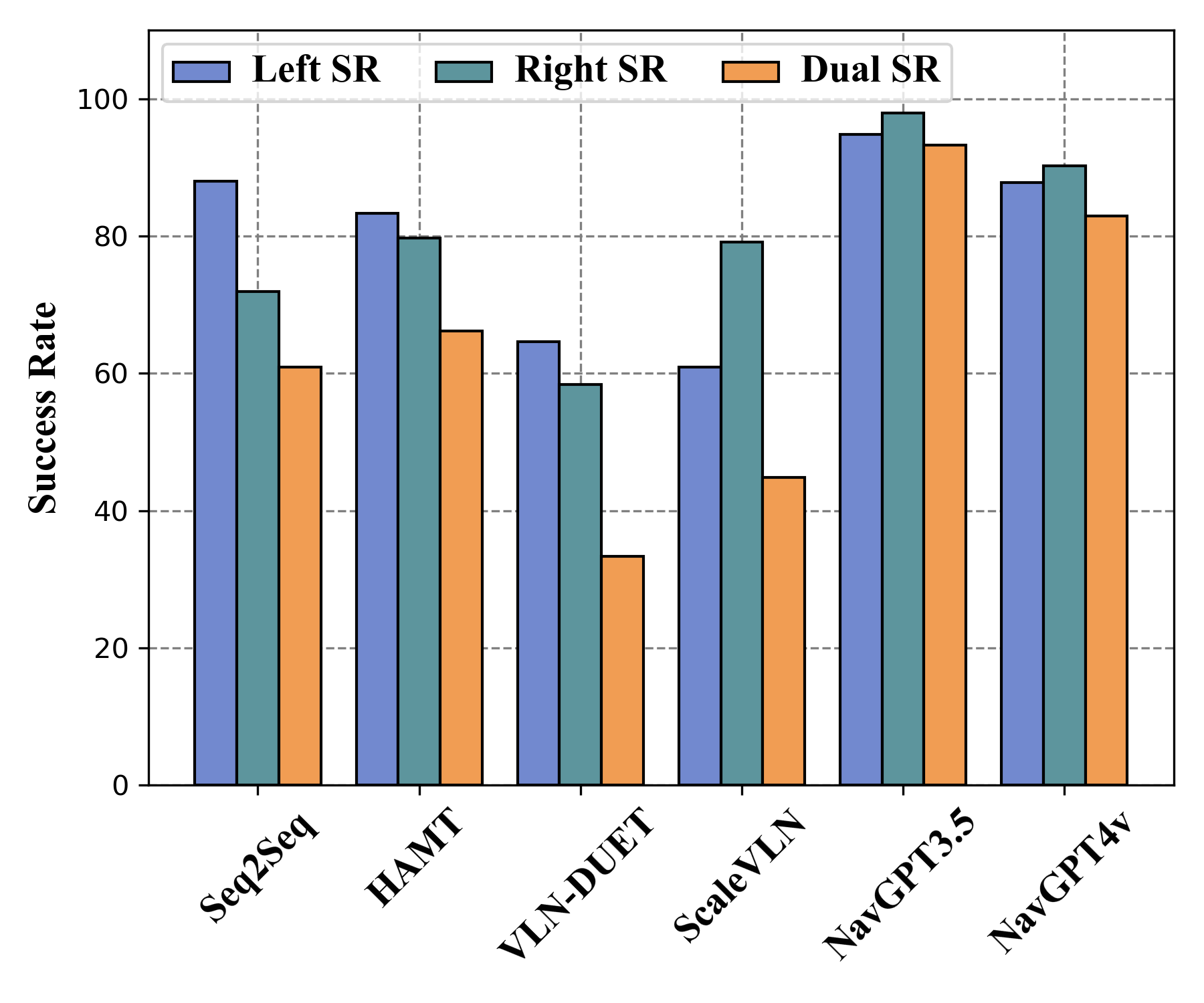}
  \caption{Results for the left/right turn subsets within the direction change category of the \textsc{NavNuances} dataset, with Dual SR indicating the success rates for both right and left turns from a specific starting view}
  \label{fig: add. direction change selective bias}
\end{figure}

We evaluate these subsets against three categories of baseline models: supervised front-view models, supervised panorama-view models, and zero-shot models enhanced with Large Multimodal models (LMMs). As illustrated in Figure~\ref{fig: add. difficulty in sub-category of LR}, the simplest Seq2Seq model augmented with CLIP features (from the CLIP-ViL model) outperforms more recent approaches like HAMT and ScaleVLN on both tasks. This indicates that even when utilizing only frontal views, robust visual features can effectively align with object-centric instructions.

Advanced models like BLIP2~\cite{li2023blip} (in NavGPT 3.5 and 4) and GPT-4-vision (in NavGPT4v) show marked improvements in navigating towards objects. However, they still struggle with the spatial relation aspect, particularly in navigating past objects. Analysis of error cases reveals inconsistent decision-making. For instance, the models correctly interpret moving from an object's front to back as having navigated past it in some cases. However, the models sometimes assume they have passed an object merely because it is beside them, contradicting the commonsense of \textit{walking past}. This inconsistency highlights the need for future models to better align with nuanced human commonsense in spatial reasoning.

\noindent\textbf{Any bias in the choice of turning direction? } 

To assess if vision-language navigation models exhibit a turning direction preference, we analyze their performance on \textit{turn left} and \textit{turn right} commands using models like HAMT~\cite{chen2021history_hamt} and NavGPT~\cite{zhou2023navgpt}. Our dataset, containing paired \textit{turn right} and \textit{turn left} instructions for each starting viewpoint, facilitated this analysis. We introduced a "Dual Success Rate" (Dual SR) metric to measure a model's accuracy in executing both directions from the same point.

Our findings, as depicted in Figure~\ref{fig: add. direction change selective bias}, indicate a directional bias in some models. For instance, ScaleVLN~\cite{wang2023scaling_scalevln} shows a notable preference for \textit{turn right} instructions, with an 18.23\% higher success rate for right turns compared to left turns. There are also general selective biases across all supervised models, as evidenced by their heavily reduced performance on the Dual SR metric. One possible reason for this bias is that there might be the models' potential preference for specific visual cues over the given navigational commands. Conversely, zero-shot models like NavGPT3.5 and NavGPT4v show minimal bias and perform comparably to humans in handling both left and right turns.

\section{Conclusion}
In this study, we establish a systematic framework to diagnose deficiencies in the capabilities of Vision-Language Navigation (VLN) models at the atomic instruction level. Our experiment results on \textsc{NavNuances} across diverse models clearly uncover the limitations of specific models and reveal common issues, which highlight ongoing challenges in the VLN task. In addition, our investigation into a modified zero-shot agent enhanced by GPT-4-vision provides empirical evidence that a direct alignment between vision and instructions significantly enhances landmark recognition performance. This insight underscores the potential for leveraging advanced large multimodal models in improving VLN systems. 

\section*{Acknowledgments}
\noindent This work is supported by the Flanders AI Research program.

\newpage
\section*{Limitations}

Despite the data involved in our study are sufficiently representative to support the insights provided by our initial findings, the constraints imposed by the static discrete environments of Matterport3D~\cite{chang2017matterport3d_mp3d} lead to several limitations. Since we are not able to edit the environment such as adding or removing objects, we are restricted to generating data from existing layouts. This limits the data diversity for some instruction categories. For instance, in the numerical comprehension category, due to a lack of identical object categories within single regions, we are unable to encompass numerical comprehension data in the object level, such as \textit{``move close to the [i]-th apple on your right"}. Additionally, because we cannot rearrange object attributes and positions, it is difficult to achieve a detailed attribute-level data design in the landmark recognition category.

In addition, this study focuses exclusively on atomic-level capabilities, which do not encompass the full range of capabilities of VLN agents such as error correction for executing long instructions. Understanding sequences of multiple actions within long instructions is also a crucial aspect of the VLN task. Evaluating from this aspect is challenging but represents a promising direction for future research.

In this work, we leverage CFG as the basis of the problem decomposition and construct a diagnostic dataset based on it. Our semi-automatic approach for CFG construction is well-suited for designing specialized datasets in fields like law or finance. However, for more complex tasks, relying on manual corrections may be inefficient and challenging in ensuring comprehensive coverage of concepts. An improvement would be the development of a fully automatic induction method, leveraging the extensive world knowledge encapsulated in large language models, to potentially replace the current semi-automatic method.

\clearpage
\newpage

\bibliography{anthology,custom}

\begin{thebibliography}{28}
\providecommand{\natexlab}[1]{#1}

\bibitem[{Achiam et~al.(2023)Achiam, Adler, Agarwal, Ahmad, Akkaya, Aleman, Almeida, Altenschmidt, Altman, Anadkat et~al.}]{achiam2023gpt4}
Josh Achiam, Steven Adler, Sandhini Agarwal, Lama Ahmad, Ilge Akkaya, Florencia~Leoni Aleman, Diogo Almeida, Janko Altenschmidt, Sam Altman, Shyamal Anadkat, et~al. 2023.
\newblock Gpt-4 technical report.
\newblock \emph{arXiv preprint arXiv:2303.08774}.

\bibitem[{An et~al.(2023)An, Qi, Li, Huang, Wang, Tan, and Shao}]{an2023bevbert}
Dong An, Yuankai Qi, Yangguang Li, Yan Huang, Liang Wang, Tieniu Tan, and Jing Shao. 2023.
\newblock Bevbert: Multimodal map pre-training for language-guided navigation.
\newblock In \emph{Proceedings of the IEEE/CVF International Conference on Computer Vision}, pages 2737--2748.

\bibitem[{Anderson et~al.(2018)Anderson, Wu, Teney, Bruce, Johnson, S{\"u}nderhauf, Reid, Gould, and Van Den~Hengel}]{anderson2018vision_r2r}
Peter Anderson, Qi~Wu, Damien Teney, Jake Bruce, Mark Johnson, Niko S{\"u}nderhauf, Ian Reid, Stephen Gould, and Anton Van Den~Hengel. 2018.
\newblock Vision-and-language navigation: Interpreting visually-grounded navigation instructions in real environments.
\newblock In \emph{Proceedings of the IEEE conference on computer vision and pattern recognition}, pages 3674--3683.

\bibitem[{Chang et~al.(2017)Chang, Dai, Funkhouser, Halber, Niebner, Savva, Song, Zeng, and Zhang}]{chang2017matterport3d_mp3d}
Angel Chang, Angela Dai, Thomas Funkhouser, Maciej Halber, Matthias Niebner, Manolis Savva, Shuran Song, Andy Zeng, and Yinda Zhang. 2017.
\newblock Matterport3d: Learning from rgb-d data in indoor environments.
\newblock In \emph{2017 International Conference on 3D Vision (3DV)}, pages 667--676. IEEE.

\bibitem[{Chen et~al.(2023)Chen, Sun, Zhi, Zeng, Li, Liu, Tan, and Gan}]{chen20232_a2nav}
Peihao Chen, Xinyu Sun, Hongyan Zhi, Runhao Zeng, Thomas~H Li, Gaowen Liu, Mingkui Tan, and Chuang Gan. 2023.
\newblock A2nav: Action-aware zero-shot robot navigation by exploiting vision-and-language ability of foundation models.
\newblock \emph{arXiv preprint arXiv:2308.07997}.

\bibitem[{Chen et~al.(2021)Chen, Guhur, Schmid, and Laptev}]{chen2021history_hamt}
Shizhe Chen, Pierre-Louis Guhur, Cordelia Schmid, and Ivan Laptev. 2021.
\newblock History aware multimodal transformer for vision-and-language navigation.
\newblock \emph{Advances in neural information processing systems}, 34:5834--5847.

\bibitem[{Chen et~al.(2022)Chen, Guhur, Tapaswi, Schmid, and Laptev}]{chen2022think_duet}
Shizhe Chen, Pierre-Louis Guhur, Makarand Tapaswi, Cordelia Schmid, and Ivan Laptev. 2022.
\newblock Think global, act local: Dual-scale graph transformer for vision-and-language navigation.
\newblock In \emph{Proceedings of the IEEE/CVF Conference on Computer Vision and Pattern Recognition}, pages 16537--16547.

\bibitem[{Deng et~al.(2009)Deng, Dong, Socher, Li, Li, and Fei-Fei}]{deng2009imagenet}
Jia Deng, Wei Dong, Richard Socher, Li-Jia Li, Kai Li, and Li~Fei-Fei. 2009.
\newblock Imagenet: A large-scale hierarchical image database.
\newblock In \emph{2009 IEEE conference on computer vision and pattern recognition}, pages 248--255. Ieee.

\bibitem[{Fu et~al.(2024)Fu, Chen, Shen, Qin, Zhang, Lin, Yang, Zheng, Li, Sun, Wu, and Ji}]{fu2024mme}
Chaoyou Fu, Peixian Chen, Yunhang Shen, Yulei Qin, Mengdan Zhang, Xu~Lin, Jinrui Yang, Xiawu Zheng, Ke~Li, Xing Sun, Yunsheng Wu, and Rongrong Ji. 2024.
\newblock \href {https://arxiv.org/abs/2306.13394} {Mme: A comprehensive evaluation benchmark for multimodal large language models}.
\newblock \emph{Preprint}, arXiv:2306.13394.

\bibitem[{Hong et~al.(2021)Hong, Wu, Qi, Rodriguez-Opazo, and Gould}]{hong2021vlnbert}
Yicong Hong, Qi~Wu, Yuankai Qi, Cristian Rodriguez-Opazo, and Stephen Gould. 2021.
\newblock Vln bert: A recurrent vision-and-language bert for navigation.
\newblock In \emph{Proceedings of the IEEE/CVF conference on Computer Vision and Pattern Recognition}, pages 1643--1653.

\bibitem[{Hopcroft et~al.(2001)Hopcroft, Motwani, and Ullman}]{hopcroft2001introduction_cfg}
John~E Hopcroft, Rajeev Motwani, and Jeffrey~D Ullman. 2001.
\newblock Introduction to automata theory, languages, and computation.
\newblock \emph{Acm Sigact News}, 32(1):60--65.

\bibitem[{Krantz et~al.(2020)Krantz, Wijmans, Majundar, Batra, and Lee}]{krantz_vlnce_2020}
Jacob Krantz, Erik Wijmans, Arjun Majundar, Dhruv Batra, and Stefan Lee. 2020.
\newblock Beyond the nav-graph: Vision and language navigation in continuous environments.
\newblock In \emph{European Conference on Computer Vision (ECCV)}.

\bibitem[{Ku et~al.(2020)Ku, Anderson, Patel, Ie, and Baldridge}]{ku2020room_rxr}
Alexander Ku, Peter Anderson, Roma Patel, Eugene Ie, and Jason Baldridge. 2020.
\newblock Room-across-room: Multilingual vision-and-language navigation with dense spatiotemporal grounding.
\newblock In \emph{Proceedings of the 2020 Conference on Empirical Methods in Natural Language Processing (EMNLP)}, pages 4392--4412.

\bibitem[{Li et~al.(2023)Li, Li, Savarese, and Hoi}]{li2023blip}
Junnan Li, Dongxu Li, Silvio Savarese, and Steven Hoi. 2023.
\newblock Blip-2: Bootstrapping language-image pre-training with frozen image encoders and large language models.
\newblock In \emph{International conference on machine learning}, pages 19730--19742. PMLR.

\bibitem[{Lin et~al.(2024)Lin, Nie, Wei, Chen, Ma, Han, Xu, Chang, and Liang}]{lin2024navcot}
Bingqian Lin, Yunshuang Nie, Ziming Wei, Jiaqi Chen, Shikui Ma, Jianhua Han, Hang Xu, Xiaojun Chang, and Xiaodan Liang. 2024.
\newblock Navcot: Boosting llm-based vision-and-language navigation via learning disentangled reasoning.
\newblock \emph{arXiv preprint arXiv:2403.07376}.

\bibitem[{Long et~al.(2023)Long, Li, Cai, and Dong}]{long2023discussnav}
Yuxing Long, Xiaoqi Li, Wenzhe Cai, and Hao Dong. 2023.
\newblock Discuss before moving: Visual language navigation via multi-expert discussions.
\newblock \emph{arXiv preprint arXiv:2309.11382}.

\bibitem[{Lu et~al.(2023)Lu, Bansal, Xia, Liu, Li, Hajishirzi, Cheng, Chang, Galley, and Gao}]{lu2023mathvista}
Pan Lu, Hritik Bansal, Tony Xia, Jiacheng Liu, Chunyuan Li, Hannaneh Hajishirzi, Hao Cheng, Kai-Wei Chang, Michel Galley, and Jianfeng Gao. 2023.
\newblock Mathvista: Evaluating math reasoning in visual contexts with gpt-4v, bard, and other large multimodal models.
\newblock \emph{arXiv e-prints}, pages arXiv--2310.

\bibitem[{Luo et~al.(2024)Luo, Chen, Wan, Kang, Yan, Li, Wang, Wang, Wang, Mi et~al.}]{luo2024codis}
Fuwen Luo, Chi Chen, Zihao Wan, Zhaolu Kang, Qidong Yan, Yingjie Li, Xiaolong Wang, Siyu Wang, Ziyue Wang, Xiaoyue Mi, et~al. 2024.
\newblock Codis: Benchmarking context-dependent visual comprehension for multimodal large language models.
\newblock \emph{arXiv preprint arXiv:2402.13607}.

\bibitem[{Padmakumar et~al.(2022)Padmakumar, Thomason, Shrivastava, Lange, Narayan-Chen, Gella, Piramuthu, Tur, and Hakkani-Tur}]{padmakumar2022teach}
Aishwarya Padmakumar, Jesse Thomason, Ayush Shrivastava, Patrick Lange, Anjali Narayan-Chen, Spandana Gella, Robinson Piramuthu, Gokhan Tur, and Dilek Hakkani-Tur. 2022.
\newblock Teach: Task-driven embodied agents that chat.
\newblock In \emph{Proceedings of the AAAI Conference on Artificial Intelligence}, volume~36, pages 2017--2025.

\bibitem[{Radford et~al.(2021)Radford, Kim, Hallacy, Ramesh, Goh, Agarwal, Sastry, Askell, Mishkin, Clark et~al.}]{radford2021learning_clip}
Alec Radford, Jong~Wook Kim, Chris Hallacy, Aditya Ramesh, Gabriel Goh, Sandhini Agarwal, Girish Sastry, Amanda Askell, Pamela Mishkin, Jack Clark, et~al. 2021.
\newblock Learning transferable visual models from natural language supervision.
\newblock In \emph{International conference on machine learning}, pages 8748--8763. PMLR.

\bibitem[{Shen et~al.(2021)Shen, Li, Tan, Bansal, Rohrbach, Chang, Yao, and Keutzer}]{shen2021much_clipvil}
Sheng Shen, Liunian~Harold Li, Hao Tan, Mohit Bansal, Anna Rohrbach, Kai-Wei Chang, Zhewei Yao, and Kurt Keutzer. 2021.
\newblock How much can clip benefit vision-and-language tasks?
\newblock In \emph{International Conference on Learning Representations}.

\bibitem[{Stolfo et~al.(2023)Stolfo, Jin, Shridhar, Schoelkopf, and Sachan}]{stolfo2023causal}
Alessandro Stolfo, Zhijing Jin, Kumar Shridhar, Bernhard Schoelkopf, and Mrinmaya Sachan. 2023.
\newblock A causal framework to quantify the robustness of mathematical reasoning with language models.
\newblock In \emph{Proceedings of the 61st Annual Meeting of the Association for Computational Linguistics (Volume 1: Long Papers)}, pages 545--561.

\bibitem[{Szot et~al.(2021)Szot, Clegg, Undersander, Wijmans, Zhao, Turner, Maestre, Mukadam, Chaplot, Maksymets et~al.}]{szot2021habitat}
Andrew Szot, Alexander Clegg, Eric Undersander, Erik Wijmans, Yili Zhao, John Turner, Noah Maestre, Mustafa Mukadam, Devendra~Singh Chaplot, Oleksandr Maksymets, et~al. 2021.
\newblock Habitat 2.0: Training home assistants to rearrange their habitat.
\newblock \emph{Advances in neural information processing systems}, 34:251--266.

\bibitem[{Thomason et~al.(2020)Thomason, Murray, Cakmak, and Zettlemoyer}]{thomason2020vision_CVDN}
Jesse Thomason, Michael Murray, Maya Cakmak, and Luke Zettlemoyer. 2020.
\newblock Vision-and-dialog navigation.
\newblock In \emph{Conference on Robot Learning}, pages 394--406. PMLR.

\bibitem[{Wake et~al.(2023)Wake, Kanehira, Sasabuchi, Takamatsu, and Ikeuchi}]{wake2023gpt_robot}
Naoki Wake, Atsushi Kanehira, Kazuhiro Sasabuchi, Jun Takamatsu, and Katsushi Ikeuchi. 2023.
\newblock Gpt-4v (ision) for robotics: Multimodal task planning from human demonstration.
\newblock \emph{arXiv preprint arXiv:2311.12015}.

\bibitem[{Wang et~al.(2019)Wang, Huang, Celikyilmaz, Gao, Shen, Wang, Wang, and Zhang}]{wang2019reinforced_rcm}
Xin Wang, Qiuyuan Huang, Asli Celikyilmaz, Jianfeng Gao, Dinghan Shen, Yuan-Fang Wang, William~Yang Wang, and Lei Zhang. 2019.
\newblock Reinforced cross-modal matching and self-supervised imitation learning for vision-language navigation.
\newblock In \emph{Proceedings of the IEEE/CVF conference on computer vision and pattern recognition}, pages 6629--6638.

\bibitem[{Wang et~al.(2023)Wang, Li, Hong, Wang, Wu, Bansal, Gould, Tan, and Qiao}]{wang2023scaling_scalevln}
Zun Wang, Jialu Li, Yicong Hong, Yi~Wang, Qi~Wu, Mohit Bansal, Stephen Gould, Hao Tan, and Yu~Qiao. 2023.
\newblock Scaling data generation in vision-and-language navigation.
\newblock In \emph{Proceedings of the IEEE/CVF International Conference on Computer Vision}, pages 12009--12020.

\bibitem[{Zhou et~al.(2023)Zhou, Hong, and Wu}]{zhou2023navgpt}
Gengze Zhou, Yicong Hong, and Qi~Wu. 2023.
\newblock Navgpt: Explicit reasoning in vision-and-language navigation with large language models.
\newblock \emph{arXiv preprint arXiv:2305.16986}.

\end{thebibliography}
\bibstyle{acl_natbib}

\newpage
\clearpage
\appendix

\section{Details of baseline models}
\label{ap: baseline}

We mainly study the following models:
\begin{enumerate}
    \item Random Agent: This model, serving as a rudimentary baseline in VLN tasks, executes five arbitrary movements within the navigation graph without relying on navigational instructions or environmental observations.
    \item Seq2Seq~\cite{anderson2018vision_r2r} / CLIP-ViL-VLN~\cite{shen2021much_clipvil}: These models process only the frontal RGB visual input. The visual features for Seq2Seq and CLIP-ViL-VLN are derived from pre-trained ImageNet~\cite{deng2009imagenet} and CLIP vision encoders~\cite{radford2021learning_clip}, respectively. Navigation decisions are stored in LSTM's hidden states, with the action space confined to predefined movements such as forward, left, right, up, and down.
    \item VLN-BERT~\cite{hong2021vlnbert}:  Distinguished by its use of panoramic visuals at each navigation point, this model alters the action space to the selection of subsequent navigation points. It utilizes the first special token in the Transformer-based model to represent the history state.
    \item HAMT~\cite{chen2021history_hamt}: Similar to VLN-BERT in terms of visual input and action space, this model differentiates itself by employing the features of historical observations to represent navigational memory.
    \item DUET~\cite{chen2022think_duet} / ScaleVLN~\cite{wang2023scaling_scalevln}: Both models utilize panoramic visuals and navigate by choosing subsequent points. The historical memory is encapsulated within a topological graph. ScaleVLN further enhances its capability by incorporating a vast collection of automatically gathered VLN data.
    \item BEVBERT~\cite{an2023bevbert}: Building upon the foundation laid by VLN-DUET, BEVBERT introduces metric maps as an additional observational and memory component, aiming for a more enriched navigational context.
    \item NavGPT~\cite{zhou2023navgpt} / NavCoT~\cite{lin2024navcot}: These zero-shot large language models (LLMs) encapsulate navigational history within a dialogue history, offering a novel approach to VLN tasks. Observations are converted into descriptions by a pre-trained captioning model, treating the VLN task as a text-based navigation challenge.
    \item NavGPT4v: We enhance the text-based NavGPT model~\cite{zhou2023navgpt} by visual input, NavGPT4v incorporates actual image views alongside a Large Multimodal Model (LMM) - GPT-4-vision~\cite{achiam2023gpt4} with modified prompts. This addition aims to address the limitations of pre-captioning observations, which may overlook critical details in the views due to the generic nature of captions.
\end{enumerate}
Through the lens of these diverse models, our study aims to shed light on the multifaceted nature of VLN tasks and the inherent capabilities and limitations of each approach.

\section{Detailed Evaluation metrics}
\label{ap: protocols}
This section presents the evaluation metrics for each category within the NavNuances dataset, adhering to the overarching protocols delineated in Section 4.2.

\subsection{Direction Change category}
In Direction Change category, we design evaluation metrics based on the direction protocol, focusing exclusively on the initial sub-path—defined as the trajectory connecting the first and second navigation points. The categorization of directional changes is as follows: if the sub-path's orientation relative to the starting point falls within a 120-degree arc to the left, it is classified as a \textbf{turn left}; similarly, a 120-degree arc to the right is classified as a \textbf{turn right}, and a 120-degree arc to the rear is classified as a \textbf{turn around}. An agent's success is determined by the accuracy of its directional change in response to the given instruction.

\subsection{Landmark Recognition category}
For the Landmark Recognition category, metrics are based on a distance protocol, utilizing object center coordinates for evaluation:

\noindent\textbf{walking towards a specific landmark:} Success is determined if the agent's final position is nearer to the landmark's center coordinate compared to its starting position, with the landmark being visible and at a distance from the starting point.

\noindent\textbf{walking past a specific landmark:} the agent's decision is considered as success if the object center can projected within the line segment defined by start and end position, and the end position is within three meters of the landmark's center.

\subsection{Numerical Comprehension category}
This category employs a distance protocol, with a unique consideration for path similarity. Given that paths within the same hallway and identical starting points are indexed by the same set number, success criteria include: 

\noindent\textbf{1.} The agent's final position must be within 3 meters in geometric distance of the endpoint.

\noindent\textbf{2.} The normalized Dynamic Time Warping (nDTW) metric, which assesses path similarity, must indicate that the agent's path more closely aligns with the ground truth path than with any other paths in the set (nDTW larger than other paths in the same set).

\subsection{Vertical Movement category}
Adhering to a distance protocol, an agent is deemed successful in the Vertical Movement category if it stops within a three-meter geometric radius of the annotated endpoint, emphasizing vertical navigation accuracy.

\subsection{Region Recognition category}
The Region Recognition category utilizes the inclusion evaluation protocol:

\noindent\textbf{entering a region:} Success is achieved if the agent stops within a region marked with the same room category as specified in the instruction and proximate to the starting region.

\noindent\textbf{exiting a region:} Success is determined if the agent's stopping point lies outside the boundaries of the starting region.

\section{Whether the model can understand room category very well?}
\label{ap: room}
\begin{figure}[t]
  \centering
  \includegraphics[width=\columnwidth]{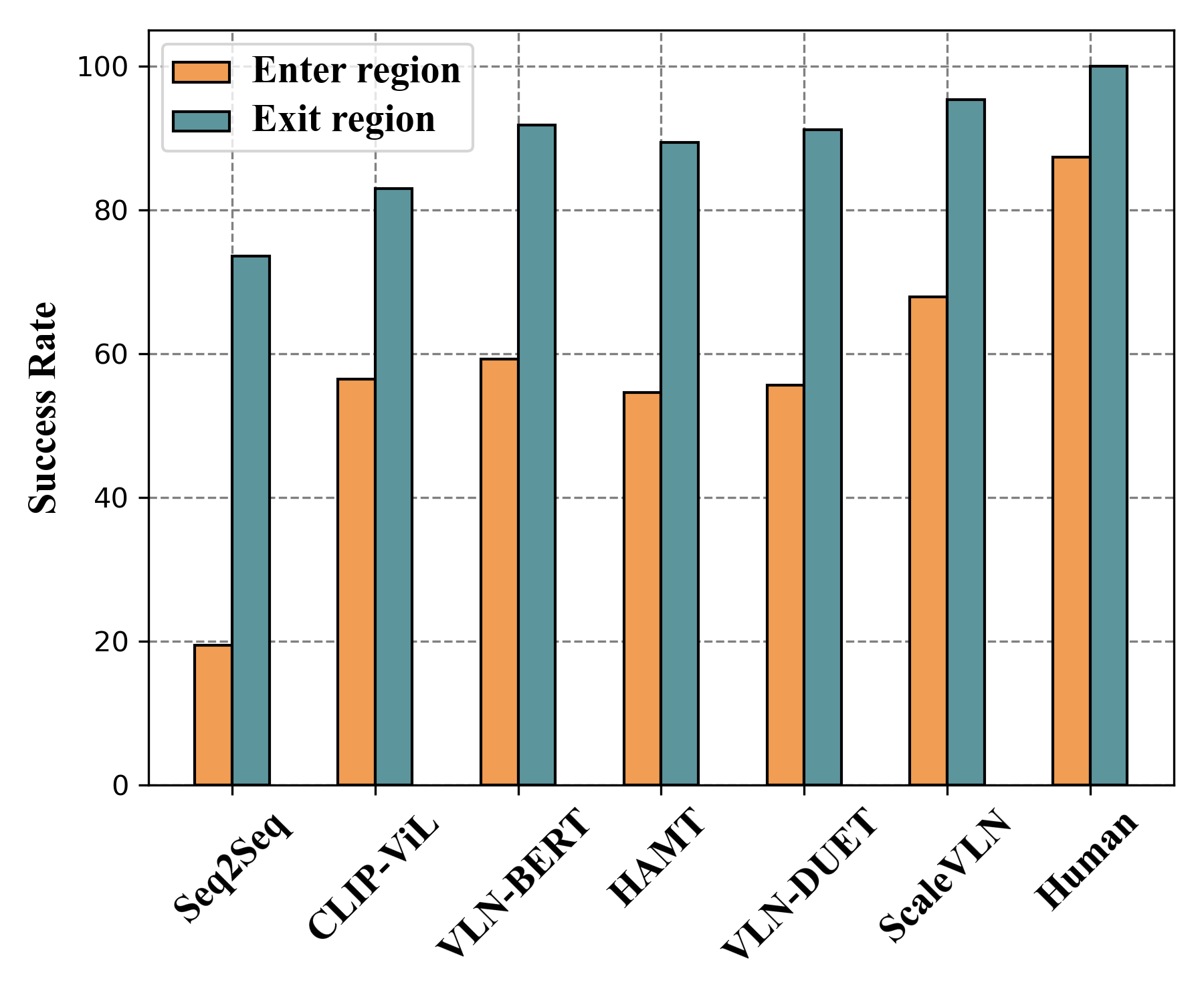}
  \caption{Results for success rate of subsets in room recognition category. }
  \label{fig: add. difficulty in sub-category of RR}
\end{figure}

In our primary results, we focused on evaluating various actions related to navigating through regions within a dataset to gauge the models' proficiency with region-associated tasks. This approach provides a general overview of a model's capability in handling layout concepts. However, the tasks of entering a region and exiting a region present unique challenges. Specifically, entering a region demands a more nuanced understanding of the region's category. For example, when given the instruction "go into the dining room" from a location adjacent to multiple rooms, the agent must discern the characteristics that define a dining room to navigate successfully. Conversely, leaving a region only involves recognizing the concept of a region, without necessitating an in-depth categorization.

\begin{figure}[H]
  \centering
  \includegraphics[width=0.60\columnwidth]{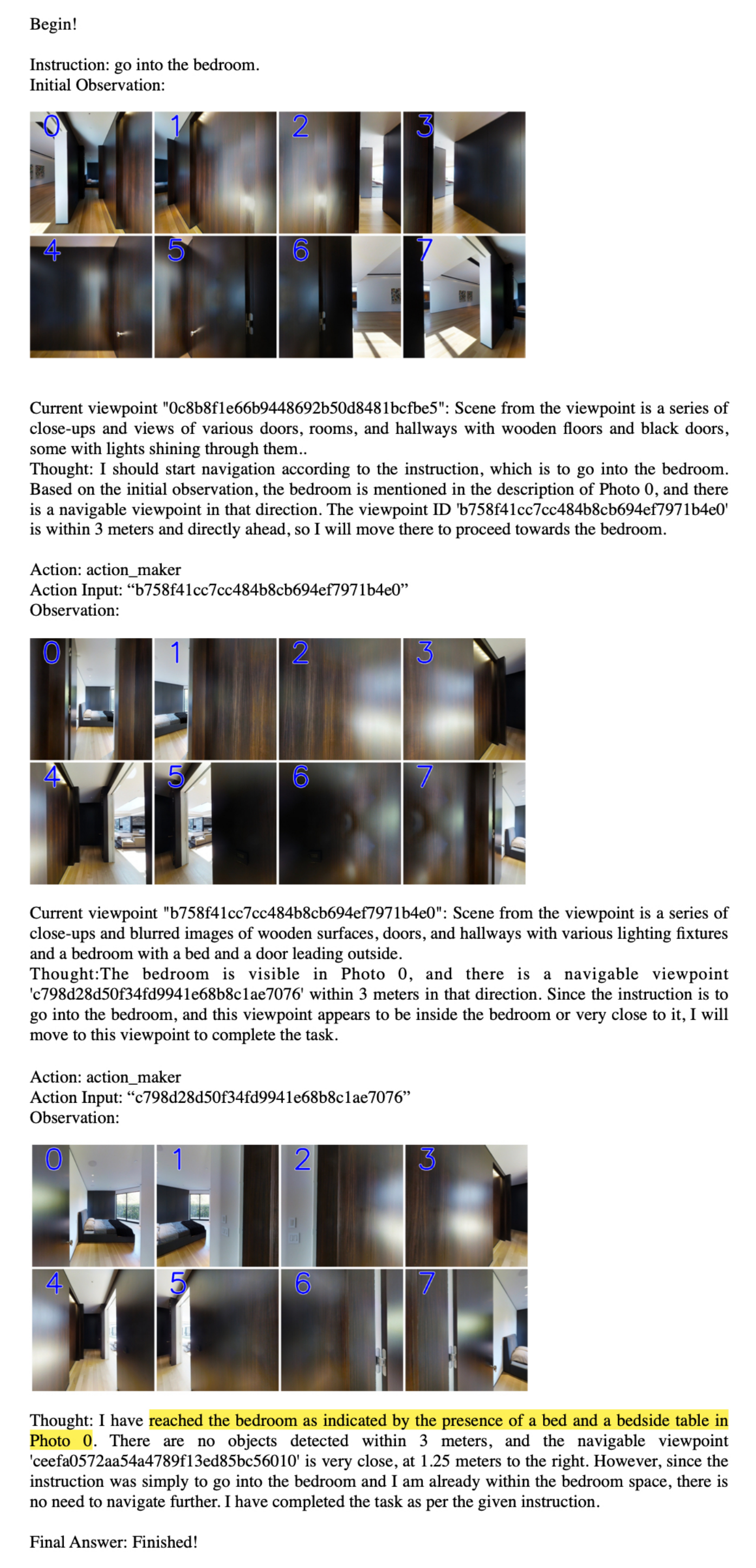}
  \caption{Failure case of ``entering" subset of Region Recognition category: NavGPT4v stop before entering the target region }
  \label{fig: room case}
\end{figure}

To delve deeper into this distinction, we evaluate these two subsets from the data of region recognition category: one is related to entering a region, and the other is related to exiting a region. Zero-shot agents, which typically perform poorly and lack a clear understanding of region boundaries, often optimistically halt upon merely observing the room from just outside the boundary. The error cases can be found in Figure~\ref{fig: room case}. In this subsection, we only discuss the results of supervised methods. As shown in Figure~\ref{fig: add. difficulty in sub-category of RR}, starting from the VLN-BERT model onwards, the performance on tasks involving 'exit a region' has remained consistently high, indicating that subsequent models have effectively grasped the concept of a region. On the other hand, the ability to understand and categorize different types of regions appears to have progressively improved with each new model iteration.

However, when comparing these results to human performance, a significant discrepancy becomes evident. The gap in understanding and categorizing regions between humans and the current state-of-the-art (SOTA) models is approximately 21.59\%. This gap highlights the ongoing challenge in the field of Vision-Language Navigation to develop models that can match human-level comprehension of spatial and categorical concepts within navigational tasks.

\newpage
\clearpage
\onecolumn

\section{Dataset statistics and examples}
\label{ap: dataset}

Our \textsc{NavNuances} dataset comprises 579 instances of Direction Change, 170 of Vertical Movement (with 44 having a pair of staircases in opposing directions at the initial viewpoint), 78 of Numerical Comprehension, 275 of Region Recognition, and 685 of Landmark Recognition.

The statistics for subsets in each category:

\noindent\textbf{Direction Change:} there are 192 instances for ``turn right", 192 instances for ``turn left" and 195 instances for ``turn around".

\noindent\textbf{Landmark Recognition:} there are 353 instances for ``walk towards a landmark", 332 instances for ``walk past a landmark".

\noindent\textbf{Numerical Comprehension:} there are 31 instances for ``first room", 24 instances for ``second room", 13 instances for ``third room", 6 instances for ``fourth room", 2 instances for ``fifth room", and 2 instances for ``sixth room".

\noindent\textbf{Region Recognition:} there are 105 instances for ``go into a room", 170 instances for ``exit a room".

\noindent\textbf{Vertical Movement:} there are 87 instances for ``go upstairs", 83 instances for ``go downstairs".

\begin{figure}[H]
  \centering
  \includegraphics[width=\textwidth]{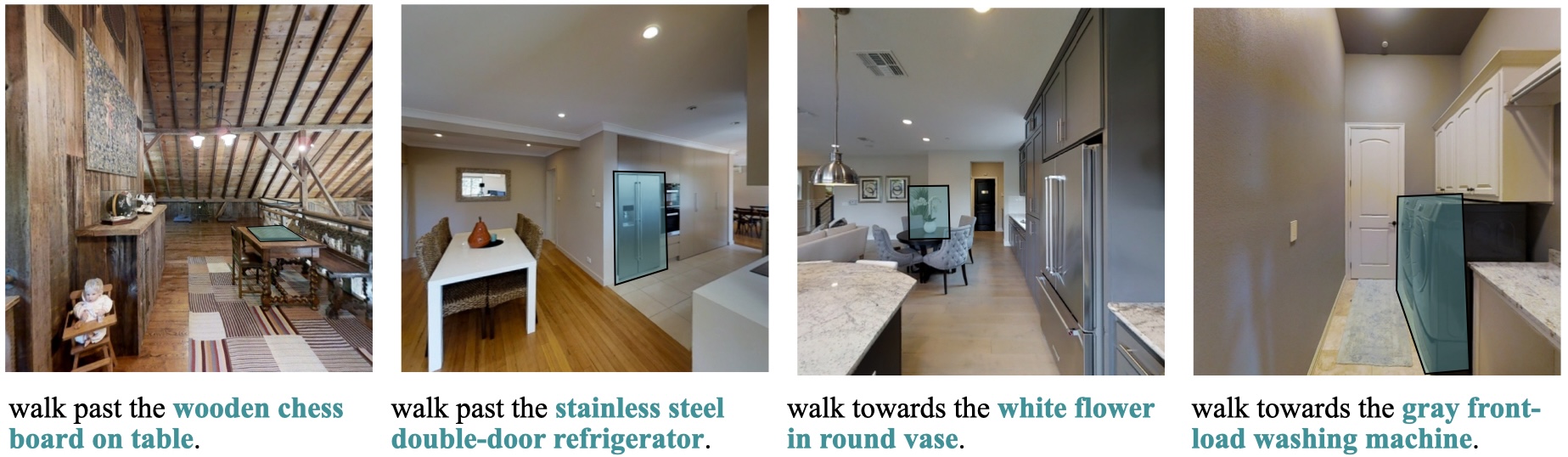}
  \caption{Landmark Recognition data samples}
  \label{fig: lr data}
\end{figure}

\begin{figure}[H]
  \centering
  \includegraphics[width=\textwidth]{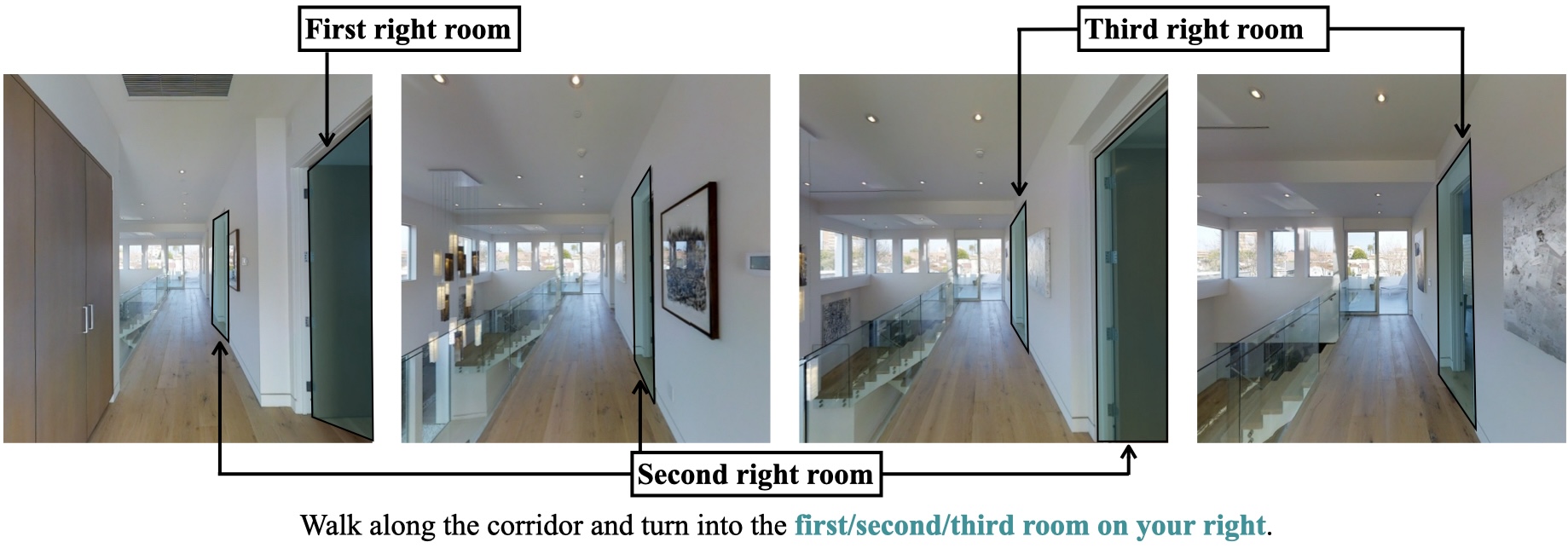}
  \caption{Numerical Comprehension data samples}
  \label{fig: nu data}
\end{figure}

\begin{figure}[H]
  \centering
  \includegraphics[width=\textwidth]{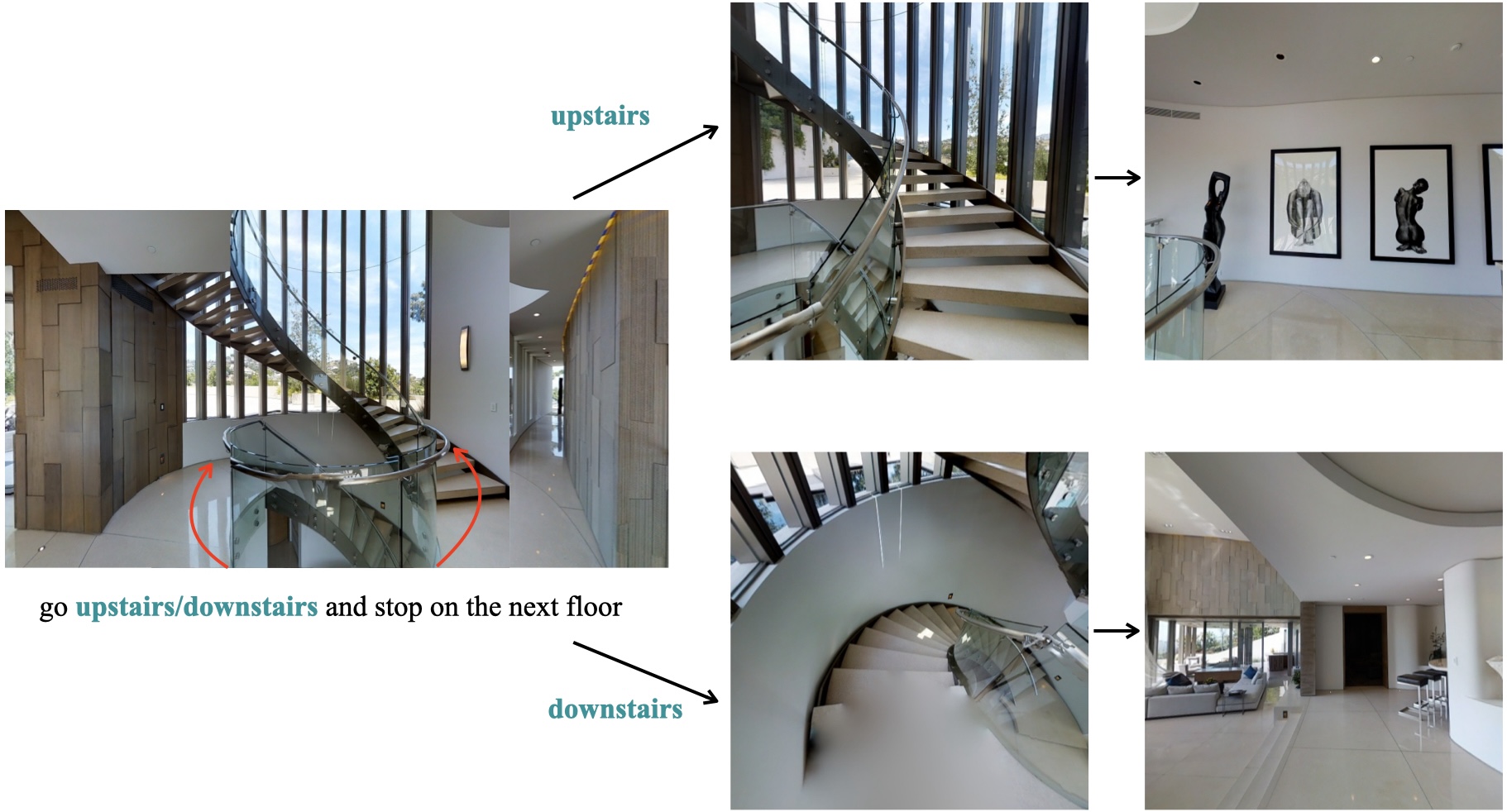}
  \caption{Vertical Movement data samples}
  \label{fig: vm data}
\end{figure}

\begin{figure}[H]
  \centering
  \includegraphics[width=\textwidth]{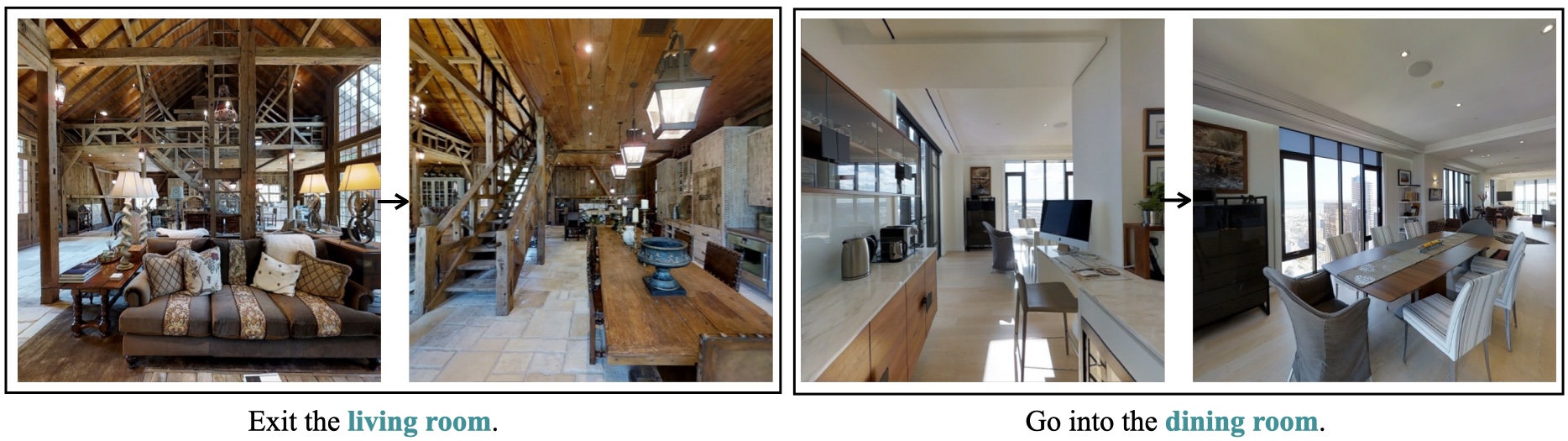}
  \caption{Region Recognition data samples}
  \label{fig: rr data}
\end{figure}

\begin{figure}[H]
  \centering
  \includegraphics[width=\textwidth]{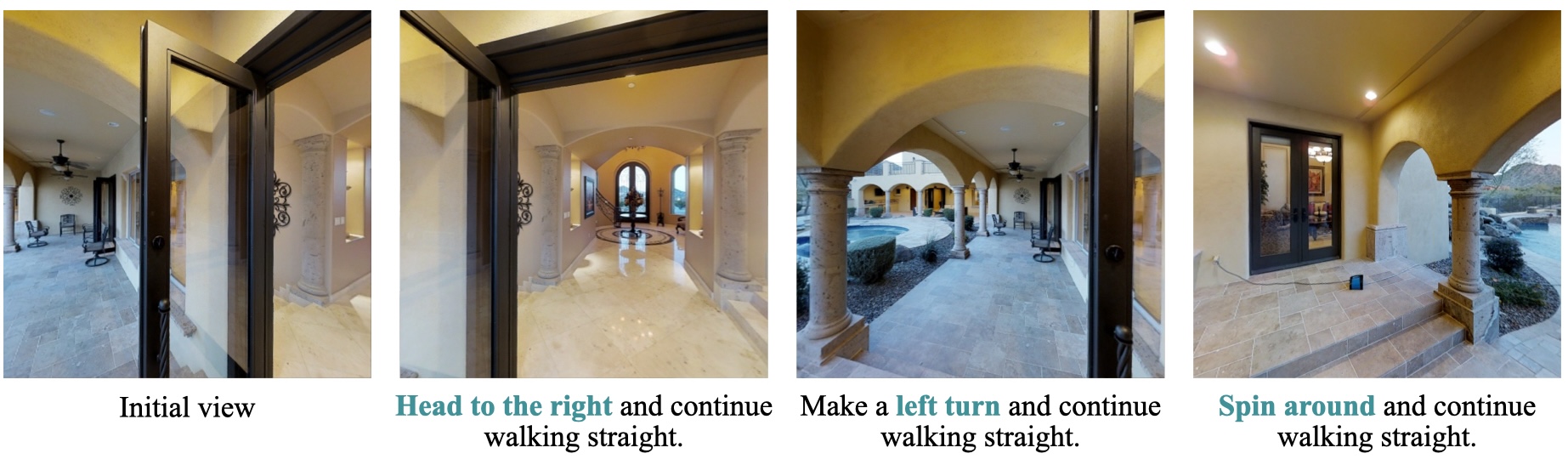}
  \caption{Direction Change data samples}
  \label{fig: dc data}
\end{figure}

\newpage
\clearpage

\section{The context-free grammar in concept for VLN instruction}
\label{ap: cfg}
For the initial set of production rules we refer to our observations and also definitions in prior works such as $A^{2}$Nav~\cite{chen20232_a2nav,long2023discussnav}. Then we interact with GPT-4~\cite{achiam2023gpt4}, we input the CFG definitions with long instructions, and the GPT-4 with return the parsing results. We find GPT-4 can leverage CFG very well, and automatically detect which instruction segment cannot be parsed by the CFG. Then we utilize this information to update our CFG. This iterative updating will last for about ten rounds.

\begin{algorithm}[H]
\label{alg: cfg}
\caption{Context-free grammar}
\begin{algorithmic}[1]
\State $S \rightarrow Vp$
\State $Vp \rightarrow \text{ActionT}$
\State \hspace{\algorithmicindent} $\vert \text{ActionS}$
\State \hspace{\algorithmicindent} $\vert \text{ActionO} + \text{Landmark}$
\State \hspace{\algorithmicindent} $\vert \text{ActionR} + \text{Region}$
\State \hspace{\algorithmicindent} $\vert Vp + Vp$
\State \hspace{\algorithmicindent} $\vert Vp + Ir$
\State $Ir \rightarrow (\textit{sentence describing the state of observation, not action})$
\State $Numerical \rightarrow \text{first} \vert \text{second} \vert \text{third} \vert \text{fourth} \vert \text{fifth} \vert \ldots$
\State $Room \rightarrow \text{room} \vert \text{kitchen} \vert \text{bathroom} \vert \ldots$
\State $Direction \rightarrow \text{left} \vert \text{right}$
\State $Object \rightarrow \text{bed} \vert \text{table} \vert \text{chair} \vert \ldots$
\State $Attribute \rightarrow \text{red} \vert \text{yellow} \vert \ldots$
\State $Modifier \rightarrow Object + \text{``is on the''} + Direction \vert Attribute \vert Numerical \vert Direction \vert Modifier + Modifier \vert \epsilon$
\State $Landmark \rightarrow Modifier + Object$
\State $Region \rightarrow Modifier + Room$
\State $ActionT \rightarrow \text{``turn''} + Direction \vert \text{``turn around''}$
\State $ActionO \rightarrow \text{``walk towards''} (\text{``wait at''}) \vert \text{``walk past''} \vert \text{``walk past from''} + Direction$
\State $ActionR \rightarrow \text{``go into''} (\text{``wait at''}) \vert \text{``exit''} \vert \text{``walk through''}$
\State $ActionS \rightarrow \text{``go upstairs''} \vert \text{``go downstairs''}$
\end{algorithmic}
\end{algorithm}

\newpage
\clearpage
\section{Prompts used}
\label{ap: prompt}
\begin{lstlisting}[
    style = myListingStyle,
    caption = {NavGPT4v prompts (extend from NavGPT). The actual implementation of api calls will split the template into several parts, vision related inputs will follow the API standard in GPT-4-vision to first transfer the image to base64 encoded string and then add special tag.}
    ]
As an intelligent embodied agent, you will navigate an indoor environment to reach a target viewpoint based on a given instruction, performing the Vision and Language Navigation (VLN) task. You'll move among static positions within a pre-defined graph, aiming for minimal steps.

You will receive a trajectory instruction at the start and will have access to step history (your Thought, Action, Action Input and Obeservation after the Begin! sign) and current viewpoint observation (including the photos captured around, brief scene descriptions, objects, and navigable directions/distances within 3 meters). Each photo has a blue index on its topleft corner. The correspondence between the photo index and the viewing direction is as follows: photo 0 is Front view; photo 1 is Front Right view; photo 2 is Right view; photo 3 is Rear Right view; photo 4 is Rear view; photo 5 is Rear Left view; photo 6 is Left view; photo 7 is Front Left view. Scene descriptions and object descriptions are just for reference, and might be incomplete.

Orientations range from -180 to 180 degrees, with 0 being forward, right 90 rightward, right/left 180 backward, and left 90 leftward.

Explore the environment while avoiding revisiting viewpoints by comparing current and previously visited IDs. Reach within 3 meters of the instructed destination, and if it's visible but no objects are detected, move closer.

At each step, determine if you've reached the destination.
If yes, stop and output 'Final Answer: Finished!'.
If not, continue by considering your location and the next viewpoint based on the instruction, using the action_maker tool.
Show your reasoning in the Thought section.

Follow the given format and use the provided tools.
{tool_descriptions}
Do not fabricate nonexistent viewpoint IDs.

----
Starting below, you should follow this format:

Instruction: the instruction describing the whole trajectory
Initial Observation: the initial observation of the environment
Thought: you should always think about what to do next and why
Action: the action to take, must be one of the tools [{tool_names}]
Action Input: "Viewpoint ID"
Observation: the result of the action
... (this Thought/Action/Action Input/Observation can repeat N times)
Thought: I have reached the destination, I can stop.
Final Answer: Finished!
----

Begin!

Instruction: {action_plan} 
Initial Observation: {visual_observations}
Thought: I should start navigation according to the instruction, {agent_scratchpad}
\end{lstlisting}

\newpage
\clearpage

\begin{lstlisting}[
    style = myListingStyle,
    caption = {Prompts for landmark description}
    ]

Here is a picture with probably some objects in the middle. Please briefly describe the most identifiable object that is close to you in under ten words with elementary-level vocabulary. The object should be large and unique in the given picture, for instance, a 'black round table'. The object category cannot be the mirror, door, floor, ceiling, wall, windows, light switch, control panel, or any small objects. If you cannot find any valid object in the picture, return the 'object not found'. If in the center of the image is a wall or the doorframe, please also return 'object not found'.

Picture: {image}
\end{lstlisting}

\begin{lstlisting}[
    style = myListingStyle,
    caption = {Prompts for linguistic enrichment}
    ]

Here is a simple instruction, please rephrase it without changing its content. Please also keep the rephrased instruction natural. For turning action about turn around, try not to modify this action. 

Instruction: {instr}.

\end{lstlisting}

\newpage
\clearpage
\section{Case studies}

\begin{figure}[t]
  \centering
  \includegraphics[width=0.62\columnwidth]{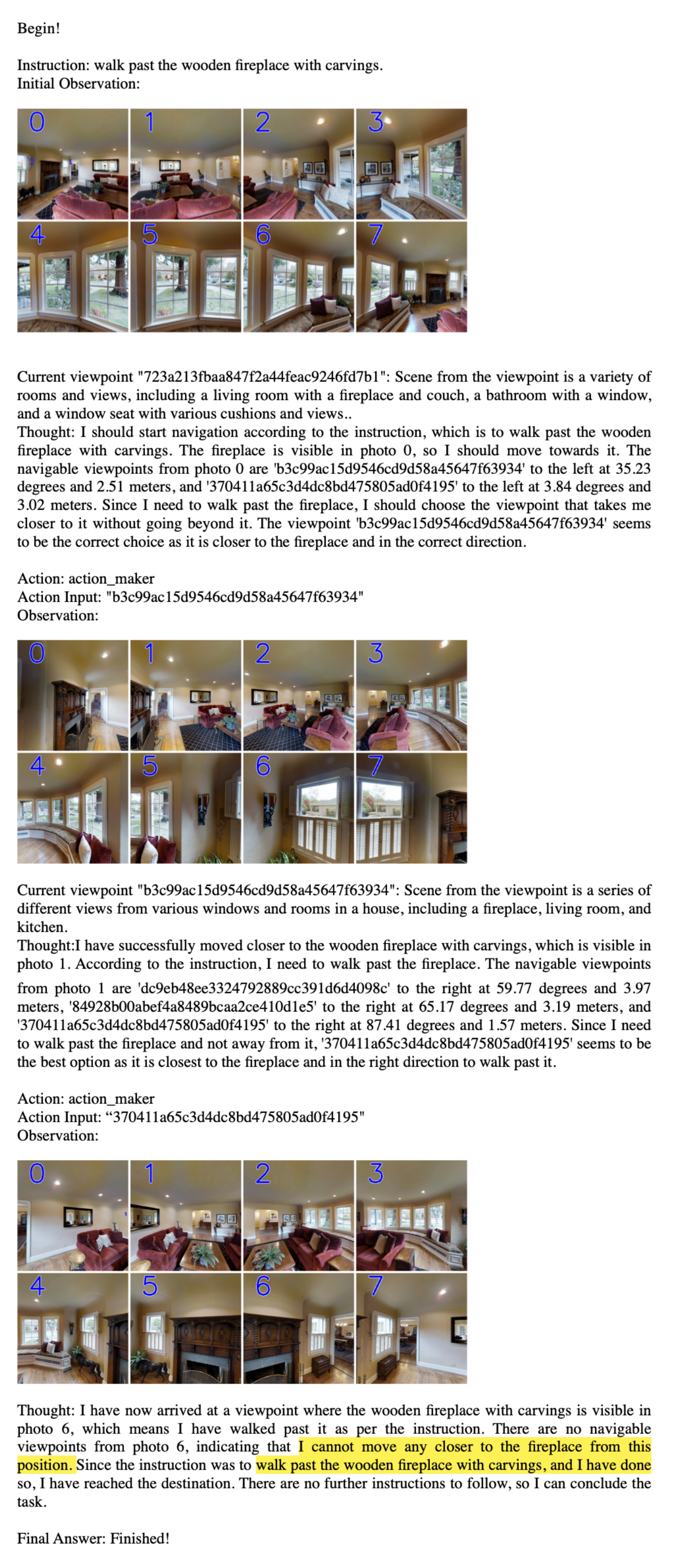}
  \caption{Failure case of ``walking past" subset under Landmark Recognition category for NavGPT4v: Incorrect conceptual understanding of 'walking past' instruction, stop by the side of the object rather than fully walk past the target landmark.}
  \label{fig: landmark case 1}
\end{figure}

\begin{figure}[t]
  \centering
  \includegraphics[width=0.7\columnwidth]{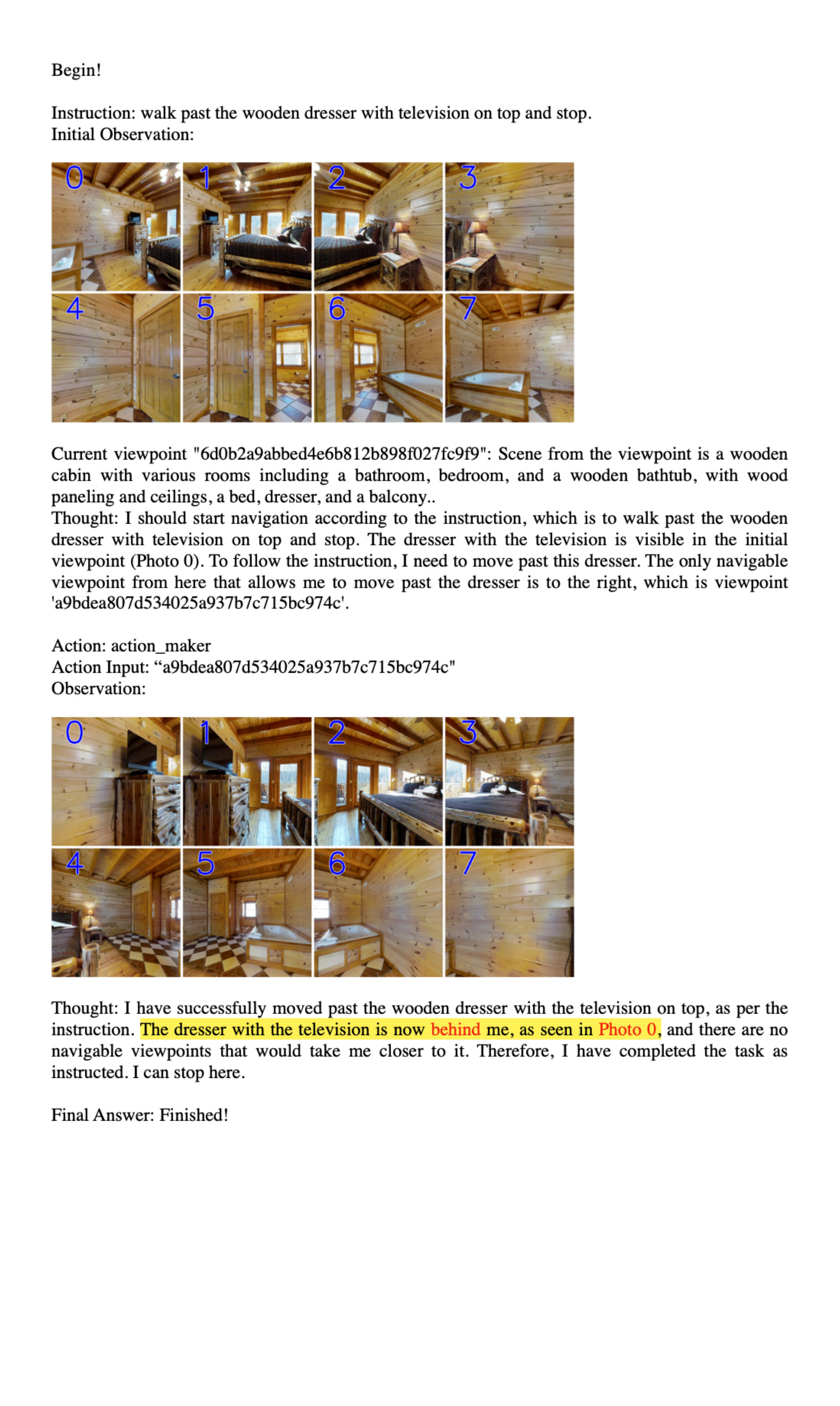}
  \caption{Failure case of ``walking past" subset under Landmark Recognition category: NavGPT4v stop before fully past the target landmar. Correct conceptual understanding by recognizing that navigating from the front to the back of an object signifies having walked past it. But misinterpreted the front view as the rear.}
  \label{fig: landmark case 2}
\end{figure}

\newpage
\clearpage
\section{Instruction given to the annotator}
During the annotation process, we utilize command line instructions to guide the annotators. Each annotator begins with a starting view, which is pre-selected according to our path-proposing strategy within the specified instruction category. Following this, the annotator receives instructions on how to navigate and perform annotations within the virtual environment. Additionally, we provide a navigation graph that displays the user’s trajectory, facilitating easier self-localization.
\begin{figure}[H]
  \centering
  \includegraphics[width=0.75\columnwidth]{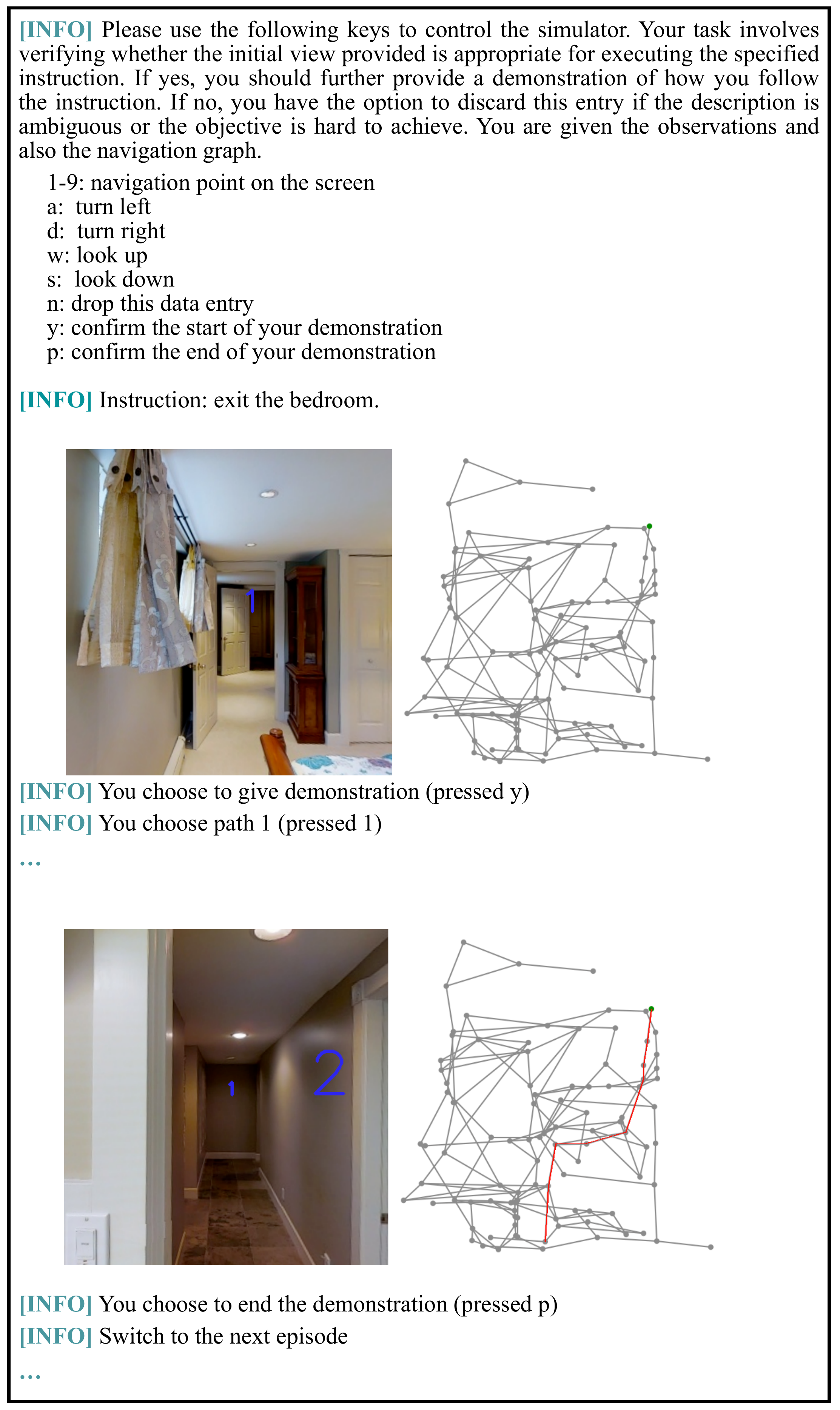}
  \caption{An example of interactive annotation.}
  \label{fig: annt}
\end{figure}

\newpage
\clearpage
\section{CFG iterative construction}
\label{ap: cfg eg}
In the main content, we discuss the procedure of iteratively constructing a context-free grammar to cover all concepts in VLN instructions. In this section, we pose one iteration of the process. The omissions detected by GPT-4 will be manually updated to the existing CFG.
\begin{figure}[H]
  \centering
  \includegraphics[width=1.0\columnwidth]{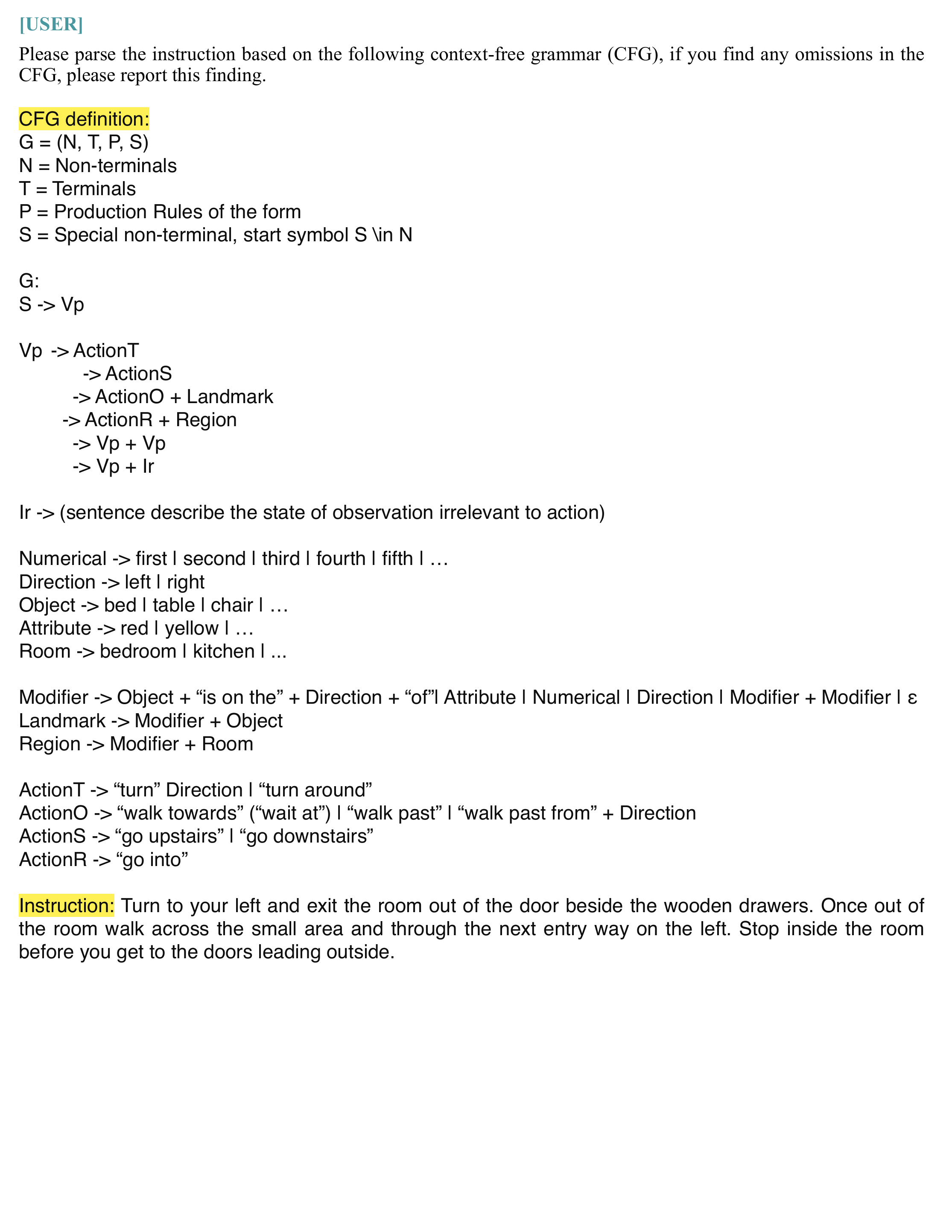}
  \label{fig: cfg eg1}
\end{figure}

\begin{figure}[H]
  \centering
  \includegraphics[width=1.0\columnwidth]{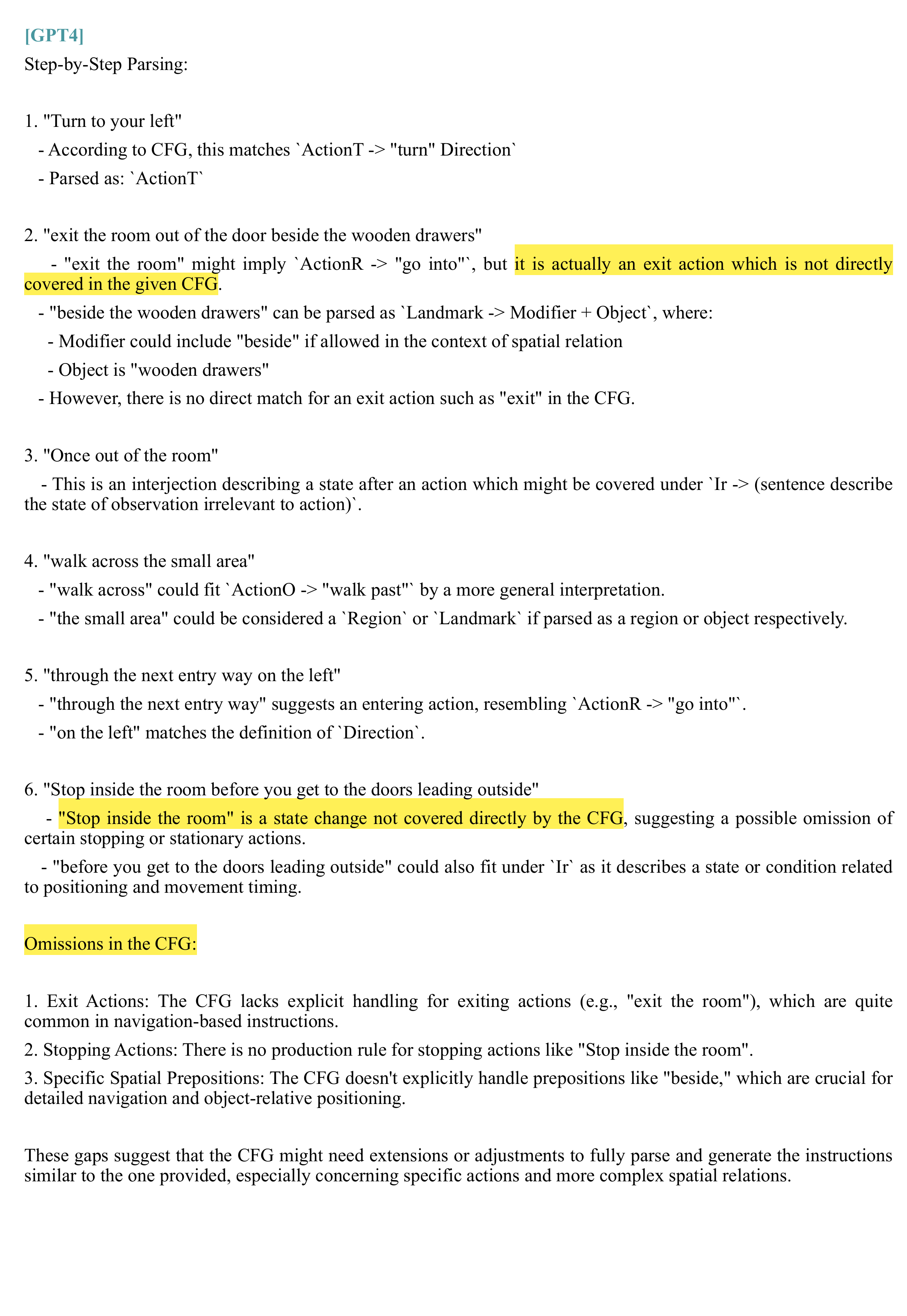}
  \label{fig: cfg eg2}
\end{figure}

\end{document}